\pdfoutput=1
\documentclass[11pt]{article}

\usepackage{ACL2023}

\usepackage{times}
\usepackage{multirow}
\usepackage{latexsym}
\usepackage{graphicx}
\usepackage{subcaption}
\usepackage{xcolor}
\usepackage{tikz}
\usepackage{amsmath}
\usepackage{enumitem}
\usepackage{booktabs}
\usepackage[most]{tcolorbox}
\usepackage{listings}
\usepackage{pgfplots}
\usepackage{placeins}
\pgfplotsset{compat=1.17}

\usepackage[T1]{fontenc}
\usepackage[utf8]{inputenc}

\usepackage{microtype}

\usepackage{inconsolata}

\newtcolorbox{inputbox}{
  colback=gray!20,
  colframe=gray!80!black,
  coltext=black,
  boxrule=0.5pt,
  arc=4pt,
  left=6pt,
  right=6pt,
  boxsep=5pt
}

\newtcolorbox{claudebox}{
  colback=blue!10,
  colframe=blue!50!black,
  coltext=black,
  boxrule=0.5pt,
  arc=4pt,
  left=6pt,
  right=6pt,
  boxsep=5pt
}

\newtcolorbox{gptbox}{
  colback=red!10,
  colframe=red!50!black,
  coltext=black,
  boxrule=0.5pt,
  arc=4pt,
  left=6pt,
  right=6pt,
  boxsep=5pt
}

\title{I am a Strange Dataset: Metalinguistic Tests for Language Models}

\author{Tristan Thrush$^{\mathsection}$, Jared Moore$^\mathsection$, Miguel Monares$^{\dagger \ddagger}$, Christopher Potts$^\mathsection$, Douwe Kiela$^{\mathsection \mathparagraph}$\\
$\mathsection$ Stanford University; $\dagger$ UC San Diego; $\ddagger$ Playtest AI; $\mathparagraph$ Contextual AI\\ \texttt{tthrush@stanford.edu} \\}

\usepackage [english]{babel}
\usepackage [autostyle, english = american]{csquotes}
\MakeOuterQuote{"}

\begin{document}
\maketitle
\begin{abstract}
Statements involving metalinguistic self-reference (``This paper has six sections.'')\ are prevalent in many domains. 
%Understanding self-reference is a key human capability. We use it in %domains from comedy to computer programming to AI research. 
Can current large language models (LLMs) handle such language?
%, and more generally, metalinguistics? 
In this paper, we present ``I am a Strange Dataset'', a new dataset for addressing this question. There are two subtasks: \emph{generation} and \emph{verification}. In generation, models continue statements like ``The penultimate word in this sentence is'' (where a correct continuation is ``is''). In  verification, models judge the truth of statements like ``The penultimate word in this sentence is sentence.''\ (false).  We also provide minimally different metalinguistic non-self-reference examples to complement the main dataset by probing for whether models can handle metalinguistic language at all.
The dataset is hand-crafted by experts and validated by non-expert annotators. We test a variety of open-source LLMs (7B to 70B parameters)
%, including Mixture of Experts (MoE) models, 
as well as closed-source LLMs through APIs. All models perform close to chance across both subtasks and even on the non-self-referential metalinguistic control data, though we find some steady improvement with model scale. GPT 4 is the only model to consistently do significantly better than chance, 
%across all of the metrics that we tested, 
and it is still only in the 60\% range, while our untrained human annotators score well in the 89--93\% range. The dataset and evaluation toolkit are available at \url{https://github.com/TristanThrush/i-am-a-strange-dataset}.
\end{abstract}

\section{Introduction}

Self-reference plays a crucial role in the way we think about mathematics~\cite{godel1931, liarparadox}, theoretical computer science~\cite{haltingproblem}, recursive programming 
%in general and self-replicating "quine" programs in particular
~\cite{geb}, understanding complex cases in hate speech detection~\cite{allan2017hard}, aptitude tests~\cite{propp1993srat}, and comedy~\cite{metamagical}. Some positions in the philosophy of mind consider self-referential capabilities to be a key aspect of higher intelligence or even consciousness~\cite{loop, consciousness}. Of course, self-reference is also pervasive in how we communicate: at least one paper you read today
%in this very collection 
is bound to contain ``In this paper''~\cite{thrush2024}. 

In this paper, we focus on metalinguistic self-reference, the complex kind of self-reference in which language is used to make claims about itself, as in ``This sentence has five words'' and ``This paper has six sections''.\footnote{Sentences like ``I am Douglas Hofstadter'' are self-referential but not metalinguistic in the sense of interest here.} Using such language involves reasoning about metalinguistic properties (counting words, naming parts of speech, etc.)\ and resolving self-reference. Humans generally have no trouble with such language, and may even enjoy its playful and sometimes paradoxical nature~\citep{geb, metamagical, loop}.

% Most people have no trouble with metalinguistics. By ``metalinguistics'', we mean language about language. We can tell that ``The following sentence has five words. I have a good cat.'' is true, whereas ``The following sentence has four words. I have a good cat.'' is false. A core subset of metalinguistics uses self-reference. We can tell that ``This sentence has five words.''\ is true whereas ``This sentence has four words.''\ is false. To interpret self-referential statements, a model must resolve the self-referent, but it also must use prerequisite metalinguistic capabilities, such as word counting.

\begin{figure}[t]
\centering
\setlength{\jot}{-2pt} % Adjust the line spacing, negative value reduces the space
\begin{subfigure}{.12\linewidth}
\raisebox{-0.4\height}{\includegraphics[width=1.2cm]{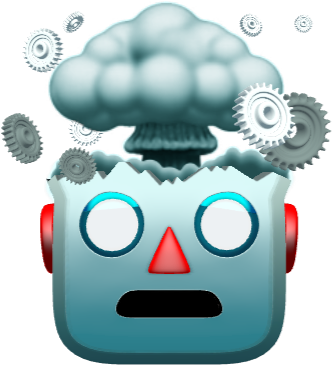}}
\end{subfigure}
\hspace{4mm}
\begin{subfigure}{.78\linewidth}
\[
P\left(\text{No.} \middle| \begin{aligned}
&\text{if someone asks whether} \\
&\text{this sentence has a capital} \\
&\text{letter, the correct answer is}
\end{aligned}\right)
\]
\end{subfigure}
\caption{An example highlighting the challenge presented by our task. All models that we tested on our dataset are close to chance-level.}
\label{fig:overview}
\end{figure}

Recently, Large Language Models (LLMs) have demonstrated striking cognitive capabilities \cite{gpt2, gpt3, chatgpt, gpt4, claude, llama2, mistral, starling}. But do they have the same mastery over metalinguistic self-reference as we do? See Figure~\ref{fig:overview} for an example of the issue that LLMs face. To help address this question, we present a new task and dataset called ``I am a Strange Dataset''. We are inspired by Douglas Hofstadter's explorations of self-reference in language~\cite{geb, metamagical, loop}, and borrow part of the name from one of his books: ``I am a Strange Loop''~\cite{loop}. 

An example in ``I am a Strange Dataset'' is comprised of two self-referential statements that begin in the same way but have different endings (Figure~\ref{fig:examples}). One is true and one is false. Crucially, the ending flips the truth value of the overall statement. There are two subtasks: \emph{generation} and \emph{verification}. In generation, the model must generate the true statement and reject the false one. In verification, models judge the truth of completed statements. To complement the main self-referential data, the dataset also contains metalinguistic non-self-reference examples. These are minimally different from the main examples and serve as controls to assess whether models can reliably handle metalinguistic statements in the absence of self-reference. In addition, all the examples in the dataset are tagged by expert annotators to further aid in error analysis.

% Our main dataset is composed of self-referential language problems, but we contrast these with another set of analogous metalinguistics problems which do not require the resolution of self-referents. We also provide a partial breakdown of metalinguistic problem categories, such as word counting, hypothetical reasoning about sentence edits, and character-level understanding. 

``I am a Strange Dataset'' is validated by non-expert annotators. As a group, they have agreement rates in the 89--93\% range, depending on which metric we use, as compared to chance rates at 50\%. This further supports the claim that metalinguistic self-reference is relatively easy for humans. LLMs, by contrast, struggle: 
``I am a Strange Dataset'' turns out to be so difficult that models are generally near chance both in generation and verification, and do not even succeed in the prerequisite metalinguistic non-self-reference case.  
% Consequently, we do not definitively decouple the challenge presented by self-reference with the challenge presented by more general metalinguistic problems. 
That said, we do find some limited evidence that GPT 4 is getting some traction on the dataset: it is significantly above chance on all tested metrics (and seems to struggle especially with the self-referential data as compared to the non-self-referential controls). However, overall, it seems safe to say that ``I am a Strange Dataset'' poses a serious challenge for even the best present-day models.

% (the only model significantly above random across all of the metrics tested) struggles more with the set that uses self-reference.

% The models that we tested are fairly close to chance, although performance rises with model scale. Non-expert humans score into the 90\% range. In the following sections, we specify the schema of our dataset, provide the results, and discuss what could be driving poor performance.

\begin{figure}[t]
  \centering
  \begin{subfigure}{.45\linewidth}
    \flushleft
    \small
    This sentence\\
    l\\
    o\\
    o\\
    k\\
    s\\
    like a letter\\
    a\\
    n\\
    d\\
    i\\
    t\\
    \textcolor{blue}{is a capital E.}
    \label{fig:instance1}
  \end{subfigure}%
  \hspace{2mm}
  \begin{subfigure}{.45\linewidth}
    \flushleft
    \small
    This sentence\\
    l\\
    o\\
    o\\
    k\\
    s\\
    like a letter\\
    a\\
    n\\
    d\\
    i\\
    t\\
    \textcolor{red}{is a capital F.}
    \label{fig:instance2}
  \end{subfigure}

  \vspace{4mm}

    \begin{lstlisting}
def a_function(a_string):
    x = "theoretically, this function"
    (*@\textcolor{blue}{a\_function("recurses infinitely")}@*)
    \end{lstlisting}

    \begin{lstlisting}
def a_function(a_string):
    x = "theoretically, this function"
    (*@\textcolor{red}{a\_function("stops eventually")}@*)
    \end{lstlisting}

  \vspace{4mm}

  \begin{subfigure}{.45\linewidth}
    \flushleft
    \small
    The first and last words of this sentence are ``The'' \textcolor{blue}{and ``respectively'', respectively.}
    \label{fig:instance5}
  \end{subfigure}%
  \hspace{2mm}
  \begin{subfigure}{.45\linewidth}
    \flushleft
    \small
    The first and last words of this sentence are ``The'' \textcolor{red}{and ``The'', respectively. ~~~~~~~~~~~~~}
    \label{fig:instance6}
  \end{subfigure}

  \caption{Examples from the dataset. Each example is comprised of a beginning and two different endings. One of the endings makes the statement true, but it would make the statement false if it referred only to the beginning. The other ending makes the statement false, but it would make the statement true if it referred only to the beginning. True endings are on the left and shown in \textcolor{blue}{blue}. False endings are on the right and shown in \textcolor{red}{red}. In the case of the code example, the true continuation is shown above the false one.}
  \label{fig:examples}

\end{figure}

\section{Related Work}

\textbf{AI Challenges}. We present our dataset as a challenge for the AI community. There are a range of AI stress tests and probes that use schemas targeting coreference resolution \cite{levesque2012winograd, sakaguchi2020winogrande}, pronoun resolution \cite{rudinger2018gender}, word order \cite{sinha2021matterslittle, thrush2022winoground, yuksekgonul2023when}, syntax \cite{linzen2016assessing, gulordava2018, gauthier-etal-2020-syntaxgym, hu-etal-2020-systematic}, and interactions between syntax and semantics \cite{kann-etal-2019-verb, thrush2020investigating}. Although the schema for these tests can be simple to describe, the knowledge required to solve the problems need not be. ``I am a Strange Dataset'' follows a simple schema that requires self-referential language, and consequently tests an array of metalinguistic capabilities. As far as we know, this is the first AI challenge dataset targeting metalinguistic self-reference, although there are metalinguistic evaluations~\cite{hu2023prompting, behzad-etal-2023-elqa} and corpuses~\cite{wilson-2012-creation-dup, kranzlein-etal-2024-curiam-corpus}.

\textbf{Self-reference}. Ideas involving self-reference have been used to boost LLM performance. LLMs can verify their own outputs either via extra passes of natural language generation \cite{weng2023large, huang2023large} or by writing code \cite{zhou2023solving}. LLMs can also enhance their own inference code to some degree \cite{zelikman2023stop}. Much of the previous work on self-reference with LLMs is about a model improving on itself or its outputs. Our dataset is not about that -- it is complementary work about a model's ability to generate and interpret metalinguistic and self-referential language. Still, a statement that refers to itself could be about a model that generated it, too: ``This sentence was generated by me, HAL9000''.

\section{I am a Strange Dataset}

In this section, we describe how the dataset is constructed and how we measure model performance.

\subsection{Dataset}

\begin{table*}[t]
\centering
\resizebox{\textwidth}{!}{
\begin{tabular}{|c|c|p{5cm}|p{4cm}|p{4cm}|}
\hline
\textbf{Tag} & \textbf{Count} & \multicolumn{3}{c|}{\textbf{Example}} \\ \cline{3-5} 
 &  & \textbf{Beginning} & \textbf{False End} & \textbf{True End} \\ \hline
Negation \& Scope & 94 & The last word you will read before the period is & not ``dog''. & actually ``dog''.\\ \hline
Numerical Operations & 62 & The number of words in this sentence is & eight. & nine. \\ \hline
Location of Element & 55 & This sentence & nothing wrong has with the word order. & something wrong has with the word order. \\ \hline
Sub-Word & 48 & Evary werd en thas sentance iz mispelled & including the words at the end. & except the words at the end. \\ \hline
Sensory & 42 & \includegraphics[width=0.35cm]{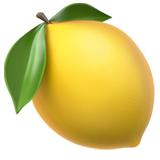} If \includegraphics[width=0.35cm]{lemon.png} you \includegraphics[width=0.35cm]{lemon.png} ate \includegraphics[width=0.35cm]{lemon.png} it, \includegraphics[width=0.35cm]{lemon.png} this \includegraphics[width=0.35cm]{lemon.png} sentence \includegraphics[width=0.35cm]{lemon.png} would \includegraphics[width=0.35cm]{lemon.png} taste \includegraphics[width=0.35cm]{lemon.png} & only \includegraphics[width=0.35cm]{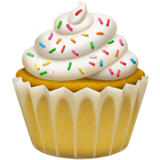} somewhat \includegraphics[width=0.35cm]{cupcake.png} sour
\includegraphics[width=0.35cm]{cupcake.png} and \includegraphics[width=0.35cm]{cupcake.png} not \includegraphics[width=0.35cm]{cupcake.png} sweet \includegraphics[width=0.35cm]{cupcake.png}. & somewhat \includegraphics[width=0.35cm]{cupcake.png} sweet \includegraphics[width=0.35cm]{cupcake.png} and \includegraphics[width=0.35cm]{cupcake.png} not \includegraphics[width=0.35cm]{cupcake.png} only \includegraphics[width=0.35cm]{cupcake.png} sour \includegraphics[width=0.35cm]{cupcake.png}. \\ \hline
Existence of Element & 31 & This sentence & lacks a verb. & has a verb. \\ \hline
Grammaticality & 25 & The author who wrote this sentence used active voice, and & only active voice is used by them. & also passive voice is used by them. \\ \hline
Multi-Channel & 24 & The penultimate word of this sentence is in the & Inglés language. & Español language. \\ \hline
Hypothetical & 24 & If you added a word here: \_ this sentence would be & eleven words. & thirteen words. \\ \hline
Question & 22 & Is there an answer that follows this question? & No. & Yes.\\ \hline
\end{tabular}
}
\caption{All of the example tags in ``I am a Strange Dataset'' sorted by count. Examples can have more than one tag.}
\label{tab:tags}
\end{table*}

We aim to test whether a language model can produce and understand self-referential statements, and has the required metalinguistic capabilities. For example, consider this incomplete statement:
\begin{quote}
\mbox{The penultimate word in this sentence is \ldots}
\end{quote}
If we did not understand metalinguistic self-reference, we might complete the sentence with the word ``sentence''. It is true that ``sentence'' is the penultimate word before adding more text, but by writing  ``sentence'', we have just changed the penultimate word! Here, a correct way to complete the statement is by inserting ``is''. Completing statements is an established task format for language models \cite{paperno2016lambada}, but as far as we know, we are the first to apply it to  metalinguistic tasks. Concretely, the schema for examples in our dataset is as follows:

\begin{itemize}[leftmargin=*,noitemsep]
    \item There is a self-referential statement which must be completed by adding text to the end.
    \item There are two candidate strings, with the same number of words, that can be used to complete the statement:
    \begin{enumerate}[noitemsep]
        \item One of the candidate strings would make the statement true if the statement refers to itself before the addition of the string, but false if it refers to itself after adding the string. An example is the answer ``sentence'' above.
        \item The other candidate string would make the statement true if the statement refers to itself after the addition of the string, but false if it refers to itself before the addition of the string. An example is the answer ``is'' above.
    \end{enumerate}
\end{itemize}

The dataset was created by four expert annotators each with several years of experience in computer science, linguistics, and/or cognitive science and all living in the United States. Each of the experts were given the schema and encouraged to be as creative as possible. Overall, the dataset is comprised of 208 examples, and split into 200 examples for the evaluation set, 3 examples for few shot prompts, and 5 examples for use in an onboarding task for non-expert human validators.

There are 10 additional examples that are completely separate from these 208 examples, which we call ``I am an Impossible Dataset''. They left even expert annotators stumped until they were given an explanation. We provide examples and GPT 4 responses in Appendix \ref{sec:appendiximpossibledataset} and leave it as inspiration for a future %(even grander!)
challenge.

\subsection{Tags}

After the dataset was created, an expert annotator came up with a set of 10 tags with which to categorize all of the examples. By using this set, we 1) ensured that there are at least 20 examples for each tag, and 2) captured aspects of the mental facilities that an expert annotator noticed when they tried solving the problems. We show the example counts for each tag in Table \ref{tab:tags}, along with representative examples from the dataset. Each example can have more than one tag. Notice that the \textbf{Sensory} tag 
 example in Table \ref{tab:tags} is also \textbf{Hypothetical}, the example for the \textbf{Existence of Element} tag is also a \textbf{Grammaticality} example, and so on. Below, we describe the knowledge categories for each tag.

\begin{enumerate}[leftmargin=*,noitemsep]
\item \textbf{Negation \& Scope}. Understanding of words such as \emph{all}, \emph{some}, \emph{most}, \emph{none}.

\item \textbf{Numerical Operations}. Arithmetic (e.g.\ multiplication, addition, counting, subtraction). It is used only if arithmetic is explicitly mentioned.

\item \textbf{Location of Element}. Where items are located in a sentence relative to everything else.

\item \textbf{Sub-Word}. Understanding of characters, morphemes, syllables, and other word components.

\item \textbf{Sensory}. Perceptual knowledge about how emojis look, how words are arranged visually, how words sound, how something might taste, etc.

\item \textbf{Existence of Element}. Whether an element is present in a statement.

\item \textbf{Grammaticality}. Knowing grammar terms.

\item \textbf{Multi-Channel}. Knowledge of at least two mediums. A medium might be Python code, English, Hebrew, C code, internet slang, etc. 

\item \textbf{Hypothetical}. Reasoning about hypotheticals.

\item \textbf{Question}. A question is involved.
\end{enumerate}
\subsection{Metrics}

We want to test whether models can generate and understand self-referential and metalinguistic statements. To this end, we present several metrics.

\subsubsection{Generation}

The primary capability that we want to test (and seemingly, the hardest) is whether 
%causal 
language models generate true self-referential statements with greater likelihood than false ones. To test for this, we take an example from the dataset and compare the losses of the continuation that makes the overall statement true versus the continuation that makes the overall statement false. If the loss of the correct continuation is lower, then the model is said to have gotten that example correct, otherwise it is incorrect. Comparing a 
%causal 
language model's surprisal of an incorrect continuation versus a correct continuation is a common method used to test for syntax-comprehension \cite{linzen2016assessing, syntaxgym} and reasoning \cite{eval-harness, mckenzie2022inverse}. Surprisal is generally proportional to the loss, $L$, of the %causal 
language model in our case. So, we define the generation score for an example's beginning $b$, true ending $e_t$, and false ending $e_f$, as given by Eq.~\ref{eq:gen}.
\begin{equation}\label{eq:gen}
        g(b,e_t, e_f)= 
    \begin{cases}
        1 & \text{if}\ L(e_t | b) < L(e_f | b) \\
        0 & \text{otherwise} \\
    \end{cases}
\end{equation}

The generation metric does not use a prompt. It is based on the loss that a model assigns to continuations, given only the beginning of a statement.

\subsubsection{Validation}

A secondary capability that we want to test is whether a %causal 
language model can at least correctly judge a given self-referential statement as true or false. To test for this, we include the self-referential statement in a prompt along with instructions that tell the model to answer whether the statement is true or not. In principle, the instructions could be anything. For our experiments, we write a standard zero-shot (ZS), few shot (FS), and chain of thought (CoT) \cite{cot} prompt. We provide the full prompts in Appendix \ref{sec:appendixmodelprompts}. 

For the ZS and FS prompts, we use the method established for the \textbf{Generation} metric above, except this time we compare the loss of \texttt{``False''} to the loss of \texttt{``True''}. Overall, the FS and ZS validation score for an example's true prompt $p_t$ (i.e.\ the true full sentence plus any instructions), and false prompt $p_f$, is given by Eq.\ \ref{eq:val}. The \textcolor{blue}{blue} parts are associated with correct model judgements and the \textcolor{red}{red} parts are associated with incorrect ones.

{\small
\begin{equation}\label{eq:val}
        v(p_t, p_f)= 
    \begin{cases}
        1 & \text{if}\ \color{blue}{L(\texttt{``True''} | p_t) < L(\texttt{``False''} | p_t)}\\
          & \text{and}\ \color{blue}{L(\texttt{``True''} | p_f) > L(\texttt{``False''} | p_f)}\\\\
          
        \frac{1}{2} & \text{if}\ \color{blue}{L(\texttt{``True''} | p_t) < L(\texttt{``False''} | p_t})\\
          & \text{and}\ \color{red}{L(\texttt{``True''} | p_f) \leq L(\texttt{``False''} | p_f)}\\\\
          
        \frac{1}{2} & \text{if}\ \color{red}{L(\texttt{``True''} | p_t) \geq L(\texttt{``False''} | p_t)}\\
          & \text{and}\ \color{blue}{L(\texttt{``True''} | p_f) > L(\texttt{``False''} | p_f)}\\\\
          
        0 & \text{if}\ \color{red}{L(\texttt{``True''} | p_t) \geq L(\texttt{``False''} | p_t)}\\
          & \text{and}\ \color{red}{L(\texttt{``True''} | p_f) \leq L(\texttt{``False''} | p_f)}\\
    \end{cases}
\end{equation}
}

We can compute the FS and ZS validation scores differently. Above, we compare the loss of \texttt{``False''} versus \texttt{``True''}, given one context at a time. We can also compare the ratios of the \texttt{``True''} and \texttt{``False''} loss, in the false versus true contexts. We call this the relative validation score because it compares a model's judgement for the truth of one sentence in an example relative to the truth of the sister sentence. This metric is given by Eq.\ \ref{eq:valrel}.

{
\begin{equation}\label{eq:valrel}
v_r(p_t, p_f)= 
\begin{cases}
    1 & \text{if}\ \frac{L(\texttt{``True''} | p_t)}{L(\texttt{``False''} | p_t)} < \frac{L(\texttt{``True''} | p_f)}{L(\texttt{``False''} | p_f)} \\
    0 & \text{otherwise}
\end{cases}
\end{equation}
}

For the CoT metric, the model is prompted to output its reasoning steps as text. We do string matching to determine the answer. Eq.\ \ref{eq:valcot} gives us the validation CoT score, where $G$ is the function that gives the model's generated text after lower-casing. Instead of string matching, we could also insert a follow-up question after the model's generation that requires a \texttt{``True''} or \texttt{``False''} and then compare log probabilities (henceforth ``logprobs''). But it is useful to have a metric in our repository that does not use logprobs, which model APIs do not always provide.

{\small
\begin{equation}\label{eq:valcot}
        v_c(p_t, p_f)= 
    \begin{cases}
        1 & \text{if}\ \color{blue}{\texttt{``true''} \in G(p_t), \texttt{``false''} \notin G(p_t)}\\
          & \text{and}\ \color{blue}{\texttt{``true''} \notin G(p_f), \texttt{``false''} \in G(p_f)}\\\\
          
        \frac{1}{2} & \text{if}\ \color{blue}{\texttt{``true''} \in G(p_t), \texttt{``false''} \notin G(p_t)}\\
          & \text{and}\ \color{red}{\neg (\texttt{``true''} \notin G(p_f), \texttt{``false''} \in G(p_f))}\\\\
          
        \frac{1}{2} & \text{if}\ \color{red}{\neg(\texttt{``true''} \in G(p_t), \texttt{``false''} \notin G(p_t))}\\
          & \text{and}\ \color{blue}{\texttt{``true''} \notin G(p_f), \texttt{``false''} \in G(p_f)}\\\\
          
        0 & \color{red}{\text{otherwise}}\\
    \end{cases}
\end{equation}
}

\subsection{Non-Self-Referential Control}

Is the self-referent part of self-referential statements (e.g.\ ``this sentence ...'') the "hard" part of metalinguistic self-reference? There are metalinguistic problem categories that are not exclusive to self-referential language: recursive phrase counting, character-level manipulation, understanding hypothetical sentence-editing scenarios, etc.

Instead of giving a language model a sentence that refers to itself, we could give it an equivalent sentence that refers to that self-referential sentence. This way, the language model would not have to know whether a self-referential sentence is true. It would only have to know whether a sentence that refers to another sentence (which happens to be self-referential) is true. This new task is still metalinguistic, but not self-referential.

It turns out that, for 97 of the sentence beginnings in ``I am a Strange Dataset'', we can replace ``this sentence'' with ``the following sentence'', and then copy the original self-referential statement below the new sentence. We can then test models for their ability to judge whether this non-self-referent version of the same statement is true. We use 2 of these 97 examples in the few shot and CoT prompts, because 2 of the examples in the original prompts cannot be turned into the non self-referent format. See Appendix \ref{sec:appendixmodelprompts} for the non-self-referent prompts. This leaves us with 95 non-self-referent examples and 95 original examples with which to compare results. GPT 4's response to a statement from the main dataset is shown in Figure \ref{fig:gpt4}, along with its response to the analagous non-self-referential control statement in Figure \ref{fig:gpt4nonselfreferential}. The responses happen to be the same in this case.

\begin{figure}[t]
\centering
% Conversation
\begin{inputbox}
\small Out out of of all all the the words words in in this this sentence sentence literally literally all all of of them them are repeated.
\end{inputbox}

\begin{gptbox}
\small \textbf{GPT 4:} Every word in the sentence is indeed repeated. So, the statement is true.
\end{gptbox}

\caption{GPT 4 misses the last words are not repeated.}
\label{fig:gpt4}

\end{figure}

\begin{figure}[t]
\centering
% Conversation

\begin{inputbox}
\small Out out of of all all the the words words in in the the following following sentence sentence literally literally all all of of them them are repeated.
\\\\
Out out of of all all the the words words in in this this sentence sentence literally literally all all of of them them are repeated.
\end{inputbox}

\begin{gptbox}
\small \textbf{GPT 4:} Every word in the sentence is indeed repeated. So, the statement is true.
\end{gptbox}
\caption{An example of GPT 4 getting a non-self-referential version of the problem from Figure \ref{fig:gpt4} wrong.}
\label{fig:gpt4nonselfreferential}
\end{figure}

\begin{table*}[t]
\centering
\resizebox{\textwidth}{!}{
\begin{tabular}{lrr|rrrrrr}
\toprule
      Model & Params & Chat & Gen$^L$ & Val ZS$^L$  & Val FS$^L$  & Val$_{rel}$ ZS$^L$ & Val$_{rel}$ FS$^L$ & Val CoT$^T$ \\
\midrule
MTurk &      - &  - &  - &      - &      \textbf{89.25} $\pm$ \textbf{3.38} &       - &       \textbf{93.00} $\pm$ \textbf{3.75} &       - \\
Random &      - &   - & 50.00 $\pm$ 0.00 &      50.00 $\pm$ 0.00 &      50.00 $\pm$ 0.00 & 50.00 $\pm$ 0.00 & 50.00 $\pm$ 0.00 &       50.00 $\pm$ 0.00 \\
\midrule
Llama 2 & 7B & N & 55.50 $\pm$ 7.00 & 50.00 $\pm$ 1.25 & 50.50 $\pm$ 2.38 & 48.50 $\pm$ 7.00 & 55.50 $\pm$ 7.00 & 5.25 $\pm$ 2.12 \\
Llama 2 & 7B & Y & 52.50 $\pm$ 7.00 & 52.25 $\pm$ 2.75 & 50.00 $\pm$ 0.75 & 52.50 $\pm$ 7.00 & 55.50 $\pm$ 6.75 & 14.00 $\pm$ 3.38 \\
Mistral 0.1 & 7B & N & 53.00 $\pm$ 6.75 & 52.25 $\pm$ 2.50 & 49.50 $\pm$ 1.50 & 56.50 $\pm$ 6.75 & 54.50 $\pm$ 7.00 & 0.00 $\pm$ 0.00 \\
Starling $\alpha$ & 7B & Y & 53.50 $\pm$ 7.00 & \textbf{54.00} $\pm$ \textbf{2.75} & 50.75 $\pm$ 1.50 & 57.00 $\pm$ 7.00 & 55.00 $\pm$ 6.75 & 35.00 $\pm$ 4.63 \\
Mistral 0.2 & 7B & Y & 52.50 $\pm$ 7.00 & 53.00 $\pm$ 4.26 & 52.25 $\pm$ 3.63 & 53.50 $\pm$ 7.00 & 53.50 $\pm$ 7.00 & 49.25 $\pm$ 4.50 \\
Llama 2 & 13B & N & 56.00 $\pm$ 7.00 & 51.50 $\pm$ 3.25 & \textbf{53.75} $\pm$ \textbf{3.50} & 50.50 $\pm$ 7.00 & \textbf{59.50} $\pm$ \textbf{6.75} & 4.50 $\pm$ 2.00 \\
Llama 2 & 13B & Y & 55.00 $\pm$ 7.00 & 52.50 $\pm$ 3.75 & 51.50 $\pm$ 2.25 & 52.50 $\pm$ 7.00 & 50.00 $\pm$ 7.00 & 9.50 $\pm$ 3.00 \\
Mixtral 0.1 & 8x7B & N & 53.50 $\pm$ 7.00 & \textbf{58.50} $\pm$ \textbf{3.75} & 51.75 $\pm$ 2.12 & 57.00 $\pm$ 7.00 & 57.00 $\pm$ 7.00 & 3.50 $\pm$ 1.88 \\
Mixtral 0.1 & 8x7B & Y & 53.50 $\pm$ 7.00 & 52.25 $\pm$ 3.75 & \textbf{53.50} $\pm$ \textbf{3.25} & 54.50 $\pm$ 7.00 & 55.50 $\pm$ 7.00 & 44.00 $\pm$ 4.75 \\
Llama 2 & 70B & N & 57.00 $\pm$ 7.00 & 53.25 $\pm$ 3.25 & \textbf{55.25} $\pm$ \textbf{2.88} & \textbf{60.00} $\pm$ \textbf{6.75} & \textbf{57.50} $\pm$ \textbf{6.75} & 2.50 $\pm$ 1.38 \\
Llama 2 & 70B & Y & 52.50 $\pm$ 7.00 & 54.25 $\pm$ 4.25 & 50.00 $\pm$ 2.00 & 56.00 $\pm$ 7.00 & \textbf{57.50} $\pm$ \textbf{6.75} & 23.50 $\pm$ 4.00 \\
Claude 2 & - & Y & - & - & - & - & - & 52.75 $\pm$ 4.00 \\
GPT 3.5 T & - & Y & - & 53.00 $\pm$ 3.00 & 53.00 $\pm$ 3.37 & 56.50 $\pm$ 7.00 & \textbf{61.00} $\pm$ \textbf{6.75} & 51.00 $\pm$ 4.63 \\
GPT 4 & - & Y & - & \textbf{59.25} $\pm$ \textbf{4.25} & \textbf{60.25} $\pm$ \textbf{4.50} & \textbf{64.50} $\pm$ \textbf{6.50} & \textbf{66.00} $\pm$ \textbf{6.50} & \textbf{66.00} $\pm$ \textbf{4.75} \\
\bottomrule
\end{tabular}
}
\caption{Comparison of models on ``I am a Strange Dataset''. Models perform fairly close to chance across all metrics. We bootstrap 95\% confidence intervals with the ``basic'' SciPy method \cite{scipy}. Metrics marked with $L$ are logprobs-based. Metrics marked with $T$ are based on generated text. We used full precision for all open source models - except the 70B models, which we used at half precision. Temperature $=0$ for all models.}
\label{tab:model_comparison}
\end{table*}

\section{Human Experiment Details}
\label{sec:humanexperiment}

To get a human baseline for our main task, we show each of the 400 self-referential statements (2 from each of the 200 examples) to at most 10 Mechanical Turk~\cite{mturk} 
workers. We separate statements from the same pair into different experiment batches. As instructions, we give the annotators the few shot prompt in Appendix \ref{sec:appendixmodelprompts}, plus an extra paragraph:

\begin{small}
\begin{verbatim}
Do not use any AI assistants such as ChatGPT to help
you; AI assistants perform very poorly at this task
and so will get many of the answers wrong. Although,
you can otherwise feel free to search online for any
information that would help you answer confidently.
For example, a few statements may contain a language
besides English. So, you can feel free to use Google
Translate. You can also search for the definitions of
words that you are unfamiliar with.
\end{verbatim}
\end{small}

To ensure a validator quality baseline, we require that all turkers are ``master'' annotators, are in the US only, have had 1000 or more previous HITs approved, have a 95\% or higher HIT approval rate, and pass a custom qualification test which we release along with the dataset. The test involves correctly answering \texttt{``True''} or \texttt{``False''} to five statements from the ``I am a Strange Dataset'' example distribution. These five statements are not used in the official evaluation set of 200 examples. The qualification test involves an acknowledgement that the annotator is familiar with basic Python programming or can search online to answer basic questions about it, because there are a few examples that require a limited understanding of code. We include a screenshot of the Mechanical Turk annotator interface in Appendix \ref{sec:appendixannotatorinterface}.

Because there are up to 10 human judgements for each statement, we get an analog to the ``loss'' for humans, and compute our metrics for humans in an analogous way to models. We can do this by computing ratios, e.g.\ $\frac{\# \text{responded true}}{10}$ and $\frac{\# \text{responded false}}{10}$. The human scores are 89.25 and 93.00 on the Val FS and Val$_{rel}$ FS metrics, as shown in Table \ref{tab:model_comparison}. Compared to the highest scoring model, the performance difference is 29 and 27, respectively. The human instructions are nearly identical to the few shot prompt, so the human responses are most comparable to the models' few shot validation responses.

\section{Results}
\label{sec:results}

\begin{table*}[t]
\centering
\begin{tabular}{c|c|c|c|c|c|}
\cline{2-6}
Tag & Question & Existence of Element & Negation \& Scope & Grammaticality & Sensory \\
Count & 62.55 & 61.01 & 57.33 & 56.32 & 55.69 \\ \cline{2-6}
\end{tabular}
\par\vspace{10pt}
\begin{tabular}{c|c|c|c|c|c|}
\cline{2-6}
Tag & Multi-Channel & Numerical Operations & Sub-Word & Hypothetical & Location of Element \\
Count & 55.16 & 51.75 & 51.52 & 48.61 & 47.72\\ \cline{2-6}
\end{tabular}
\caption{Results for all of the example tags in ``I am a Strange Dataset'' sorted by score. Scores are averaged for all models and all logprobs-based metrics (so each score here is an average from 63 scores).}
\label{tab:tags_results}
\end{table*}

\begin{table*}[t]
\centering
\resizebox{\textwidth}{!}{
\begin{tabular}{lrr|rrrrrr}
\toprule
Model & Params & Chat & $\Delta$ Gen$^L$ & $\Delta$ Val ZS$^L$ & $\Delta$ Val FS$^L$ & $\Delta$ Val ZS$^L$ (R) & $\Delta$ Val FS$^L$ (R) & $\Delta$ Val CoT$^T$ \\
\midrule
Llama 2 & 7B & N & -4.21 $\pm$ 9.47 & -1.58 $\pm$ 4.21 & 1.05 $\pm$ 3.16 & 8.42 $\pm$ 9.47 & 4.21 $\pm$ 10.53 & \textbf{30.53} $\pm$ \textbf{6.07} \\
Llama 2 & 7B & Y & -2.11 $\pm$ 10.00 & -2.11 $\pm$ 4.21 & 0.53 $\pm$ 0.79 & 1.05 $\pm$ 10.53 & -1.05 $\pm$ 12.11 & \textbf{27.37} $\pm$ \textbf{6.58} \\
Mistral 0.1 & 7B & N & -3.16 $\pm$ 10.00 & \textbf{-4.74} $\pm$ \textbf{3.68} & 1.58 $\pm$ 2.11 & 5.26 $\pm$ 10.53 & 1.05 $\pm$ 10.00 & 0.00 $\pm$ 0.00 \\
Starling $\alpha$ & 7B & Y & 0.00 $\pm$ 9.47 & -3.16 $\pm$ 3.68 & -1.58 $\pm$ 2.37 & 0.00 $\pm$ 11.58 & -6.32 $\pm$ 11.58 & 4.21 $\pm$ 8.68 \\
Mistral 0.2 & 7B & Y & -4.21 $\pm$ 10.00 & -2.63 $\pm$ 7.89 & -3.16 $\pm$ 5.79 & 2.11 $\pm$ 11.58 & 2.11 $\pm$ 11.05 & \textbf{-8.95} $\pm$ \textbf{8.68} \\
Llama 2 & 13B & N & -9.47 $\pm$ 10.00 & -0.53 $\pm$ 5.26 & 0.53 $\pm$ 5.26 & -2.11 $\pm$ 7.37 & 4.21 $\pm$ 11.58 & \textbf{23.68} $\pm$ \textbf{6.32} \\
Llama 2 & 13B & Y & -3.16 $\pm$ 10.00 & -1.05 $\pm$ 5.26 & -2.11 $\pm$ 3.42 & 2.11 $\pm$ 9.47 & 2.11 $\pm$ 11.58 & \textbf{16.32} $\pm$ \textbf{6.32} \\
Mixtral 0.1 & 8x7B & N & -1.05 $\pm$ 10.00 & -2.63 $\pm$ 4.74 & -0.53 $\pm$ 2.63 & -2.11 $\pm$ 10.53 & 6.32 $\pm$ 11.58 & -1.05 $\pm$ 2.11 \\
Mixtral 0.1 & 8x7B & Y & -5.26 $\pm$ 10.00 & 5.26 $\pm$ 6.84 & 4.74 $\pm$ 5.26 & 7.37 $\pm$ 11.58 & 3.16 $\pm$ 11.58 & \textbf{-25.26} $\pm$ \textbf{9.21} \\
Llama 2 & 70B & N & -7.37 $\pm$ 10.03 & 2.11 $\pm$ 5.79 & -3.68 $\pm$ 5.26 & 5.26 $\pm$ 10.53 & 6.32 $\pm$ 11.05 & -0.53 $\pm$ 0.79 \\
Llama 2 & 70B & Y & -1.05 $\pm$ 10.53 & -1.58 $\pm$ 5.79 & 1.58 $\pm$ 4.21 & 3.16 $\pm$ 9.47 & 5.26 $\pm$ 8.42 & \textbf{-25.79} $\pm$ \textbf{5.79} \\
Claude 2 & - & Y & - & - & - & - & - & 3.16 $\pm$ 6.84 \\
GPT 3.5 T & - & Y & - & 4.21 $\pm$ 6.84 & 3.16 $\pm$ 6.32 & -2.11 $\pm$ 11.58 & -2.11 $\pm$ 11.58 & -3.68 $\pm$ 8.95 \\
GPT 4 & - & Y & - & \textbf{12.11} $\pm$ \textbf{6.84} & 3.68 $\pm$ 6.84 & 10.53 $\pm$ 10.53 & 6.32 $\pm$ 10.53 & 1.05 $\pm$ 7.37 \\
\bottomrule
\end{tabular}

}
\caption{The difference between scores on ``I am a Strange Dataset'' when the referent is ``the following sentence'' instead of ``this sentence'' (scores for the first minus the latter). Overall, the problems that LLMs have with self-referential statements do not stem only from issues understanding the self-referential referent itself. Differences outside of the 95\% confidence interval (computed the same way as for Table \ref{tab:model_comparison}) are shown in \textbf{bold}.}
\label{tab:model_comparison_delta}
\end{table*}

\begin{figure}[t]
\centering
% Conversation

\begin{inputbox}
\small The numerical digit symbol ``1'' appears in this sentence exactly 1 plus one times.
\end{inputbox}

\begin{gptbox}
\small \textbf{GPT 4:} The text has 1 numerical digit symbol ``1'' and the word ``one'' appears once. So, the statement is true.
\end{gptbox}

\caption{Arguably, an example where GPT 4 should not have gotten points. This is an example where GPT~4 chooses the correct true/false response, but with incorrect reasoning. The ``1'' symbol appears twice.}
\label{fig:gpt4falsepositive}
\end{figure}

Table \ref{tab:model_comparison} showcases our results on a variety of open-source \cite{llama2, mistral, starling, mixtral} and closed-source~\cite{gpt3, gpt4, claude} models. Overall, the models perform close to the level of chance. The only model to achieve scores significantly above random on all metrics tested is GPT~4, and even so, the performance is well below the non-expert human scores. 

Results in this paper for Claude 2 \cite{claude}, GPT 3.5 Turbo \cite{gpt3}, and GPT 4 \cite{gpt4} were collected through the \texttt{gpt-4}, \texttt{gpt-3.5-turbo}, and \texttt{claude-2} endpoints on their respective APIs on Jan 7, 2024 (unless otherwise stated). Note that the Claude 2 API does not support logprobs, so the only metric that we report for it is the text-based CoT validation metric. The OpenAI API supports top-5 logprobs access, and it turns out that this is enough to get logprobs for \texttt{``True''} vs \texttt{``False''} in our experiments. It is not enough to get the multi-token logprobs required for the generation metric, though.

The CoT metric gives us extra insight into the limitations of models because we can see if their reasoning aligns with their final answer. In many cases, they make fairly obvious mistakes as seen in Figure \ref{fig:gpt4}. In some cases, the models choose the correct answer even though their reasoning is flawed, as seen in Figure \ref{fig:gpt4falsepositive}. 

It is important to note that there are limitations with the CoT validation metric, and the ZS and FS non-relative validation metrics too. These metrics are about judging a statement as true or false, independent of the statement with the alternative continuation. The dataset schema calls for sentences which are either true or false, but in practice, language can be ambiguous. Figure \ref{fig:gpt4falsenegative} illustrates. A fairer gauge of model capability may be to compare the continuations with each other, as the generation and relative validation metrics do.

\textbf{Tags.} We present the average logprobs-based model scores for the tags in Table \ref{tab:tags_results}. We opt to leave CoT scores out due to their high variance (many of the open-source models do not even return either \texttt{``true''} or \texttt{``false''} in their answers). Full tables for the tags are in Appendix \ref{sec:appendixtagresults}.

\textbf{Non-Self-Referential Control.} We provide the difference in scores between the non-self-referent examples and analogous original examples in Table~\ref{tab:model_comparison_delta}. The significant values in the CoT column are not particularly compelling because the open-source models are below chance regardless of whether there is self-reference. For this metric, the open-source models are extremely sensitive to prompting -- many of them do not even output \texttt{``true''} or \texttt{``false''} for every example. Although stronger models do so reliably - only 7 out of all of GPT 4's responses contain both a \texttt{``true''} and a \texttt{``false''}, or neither.

We find that our dataset is so hard due to non-self-referential metalinguistic aspects that models score around chance here too, with the exception of GPT 4. It is the only model which is strong enough to perform significantly above chance for every metric tested on the main dataset in Table \ref{tab:model_comparison}, and to have all positive values in Table \ref{tab:model_comparison_delta} for every metric tested, including logprobs-based metrics (meaning that the self-referential version was harder for it). Although, only the Val ZS value for GPT 4 is well outside of the 95\% confidence interval, and GPT 4 is also still not particularly good at the non-self-referential version. Figure \ref{fig:gpt4nonselfreferential} shows that \mbox{GPT 4} struggles with the non-self-referent version of the Figure \ref{fig:gpt4} example. There is some signal that the challenge posed by self-referents will remain as LLMs gain competence at other metalinguistic problems, but the dataset is so hard that we do not have overwhelming evidence.

\textbf{Model Scale.} If we exclude the high-variance CoT metric, we see a clear scaling trend that models with more parameters score higher on the test. See Figure \ref{fig:paramstoavgscore}. Will this trend continue? For additional discussion, see Appendix \ref{sec:appendixsupplementaldiscussion}.

\begin{figure}[t]
\centering
% Conversation

\begin{inputbox}
\small Why doesn't this question have an answer? It does, though.
\end{inputbox}

\begin{gptbox}
\small \textbf{GPT 4:} The statement is asking why it doesn't have an answer, but it does have an answer. So, the statement is false.
\end{gptbox}

\caption{This is a particularly ambiguous example that is assigned the gold label of true. The response from GPT 4 is marked as incorrect, although one could make a case that the statement is false because of the false presupposition which is corrected later in the answer. The generation and relative validation metrics are robust to these situations. For these metrics, the losses from \texttt{``It does though.''}\ versus \texttt{``It just doesn't.''}\ (the alternative continuation) are compared directly. Regardless of ambiguity about whether a statement is true outright, we can notice a higher confidence for \texttt{``It just doesn't.''}\ as the false continuation.}
\label{fig:gpt4falsenegative}
\end{figure}

\begin{figure}[t]
\centering
\resizebox{\columnwidth}{!}{
\begin{tikzpicture}
\begin{axis}[
    title={Params vs Avg logprobs-based Score},
    xlabel={Params (B)},
    ylabel={Score},
    grid=major,
    xmin=0, xmax=70,
    ymin=52, ymax=56
]
\addplot[mark=o, blue] coordinates {

(7.0, 52.94)
(13.0, 53.275)
(56.0, 54.7)
(70.0, 55.325)

};

\end{axis}
\end{tikzpicture}
}
\caption{Parameters to average ``I am a Strange Dataset'' score across all of the logprobs-based metrics. We only evaluate five 7B models, and two models for each of the other sizes, so computing confidence intervals for each point is not particularly informative. Under the null hypothesis that parameter size has no effect on score, we can compute the $p$-value for these results as a whole nonparametrically: there are 24 ways that these 4 points can be arranged and in only 1 of the ways do they all increase with the parameter count: $p = 1/24 = 0.042$.}
\label{fig:paramstoavgscore}
\end{figure}

\section{Conclusion}

A grasp of self-reference is important in a variety of domains, and is a notable aspect of human intelligence. We introduced a novel task and dataset to assess the capacity of models to generate and understand self-referential statements, and understand the prerequisite metalinguistic reasoning. All models that we tested perform fairly close to the level of chance. GPT 4 is the only model to score significantly above chance on all of the metrics tested, and still it is not by much. The poor performance may be indicative of a larger issue about the limitations of even today's best causal language models. Even though the task is straightforward for people, we find evidence that scale beyond 70B parameters may be needed for the emergence of comparable performance from models.

Our results indicate that this dataset is hard not only due to the self-referent part of a self-referential statement. The challenge also comes from other metalinguistic aspects, such as recursively applying arithmetic operations on sentences. Still, there is some limited evidence that GPT 4 struggles more with self-referential metalinguistic problems than analogous non-self-referential problems.

\section{Dataset Release Strategy}

We release the dataset on GitHub. at \url{https://github.com/TristanThrush/i-am-a-strange-dataset}. The data is encrypted, but the decryption script is provided. Our goal is not to hide the dataset from people, but to hide the dataset from any processes that scrape training data from the web. We encourage the rest of the community to take up this practice when releasing evaluation datasets in a public repository.

\section{Limitations}

It is possible that the self-reference aspect of ``I am a Strange Dataset'' will turn out to be the bottleneck for many models, but it is also true that models are largely failing at the purely metalinguistic aspect. Although the schema targets metalinguistic self-reference, it is difficult to make a specific claim about why models fail without running more experiments and without waiting until models become more competent.

\section{Ethical Considerations}

We aimed to pay crowdworkers 15 USD hourly based on an estimated task completion time.

%\section*{Acknowledgements}

%We are grateful to Rohan Pandey for contributing a few examples to the dataset and to Aryaman Arora for giving us the idea to cite this paper within itself. The emoji artwork is from Apple.

\bibliography{anthology,custom}

\begin{thebibliography}{46}
\expandafter\ifx\csname natexlab\endcsname\relax\def\natexlab#1{#1}\fi

\bibitem[{Allan(2017)}]{allan2017hard}
Richard Allan. 2017.
\newblock \href {https://about.fb.com/news/2017/06/hard-questions-hate-speech/} {Hard questions: Who should decide what is hate speech in an online global community?}

\bibitem[{Amazon(Retrieved 2023)}]{mturk}
Amazon. Retrieved 2023.
\newblock \href {https://www.mturk.com/} {Mechanical turk}.

\bibitem[{Anthropic(2023)}]{claude}
Anthropic. 2023.
\newblock \href {https://www.anthropic.com/index/introducing-claude} {Claude}.

\bibitem[{Baars(1993)}]{consciousness}
Bernard~J. Baars. 1993.
\newblock \emph{A cognitive theory of consciousness}.
\newblock Cambridge University Press.

\bibitem[{Behzad et~al.(2023)Behzad, Sakaguchi, Schneider, and Zeldes}]{behzad-etal-2023-elqa}
Shabnam Behzad, Keisuke Sakaguchi, Nathan Schneider, and Amir Zeldes. 2023.
\newblock {ELQA}: A corpus of metalinguistic questions and answers about {E}nglish.
\newblock In \emph{ACL}.

\bibitem[{Brown et~al.(2020)Brown, Mann, Ryder, Subbiah, Kaplan, Dhariwal, Neelakantan, Shyam, Sastry, Askell, Agarwal, Herbert-Voss, Krueger, Henighan, Child, Ramesh, Ziegler, Wu, Winter, Hesse, Chen, Sigler, Litwin, Gray, Chess, Clark, Berner, McCandlish, Radford, Sutskever, and Amodei}]{gpt3}
Tom~B. Brown, Benjamin Mann, Nick Ryder, Melanie Subbiah, Jared Kaplan, Prafulla Dhariwal, Arvind Neelakantan, Pranav Shyam, Girish Sastry, Amanda Askell, Sandhini Agarwal, Ariel Herbert-Voss, Gretchen Krueger, Tom Henighan, Rewon Child, Aditya Ramesh, Daniel~M. Ziegler, Jeffrey Wu, Clemens Winter, Christopher Hesse, Mark Chen, Eric Sigler, Mateusz Litwin, Scott Gray, Benjamin Chess, Jack Clark, Christopher Berner, Sam McCandlish, Alec Radford, Ilya Sutskever, and Dario Amodei. 2020.
\newblock Language models are few-shot learners.
\newblock \emph{arXiv}.

\bibitem[{Church(1936)}]{haltingproblem}
Alonzo Church. 1936.
\newblock An unsolvable problem of elementary number theory.
\newblock \emph{American Journal of Mathematics}.

\bibitem[{Gao et~al.(2021)Gao, Tow, Biderman, Black, DiPofi, Foster, Golding, Hsu, McDonell, Muennighoff, Phang, Reynolds, Tang, Thite, Wang, Wang, and Zou}]{eval-harness}
Leo Gao, Jonathan Tow, Stella Biderman, Sid Black, Anthony DiPofi, Charles Foster, Laurence Golding, Jeffrey Hsu, Kyle McDonell, Niklas Muennighoff, Jason Phang, Laria Reynolds, Eric Tang, Anish Thite, Ben Wang, Kevin Wang, and Andy Zou. 2021.
\newblock \href {https://doi.org/10.5281/zenodo.5371628} {A framework for few-shot language model evaluation}.

\bibitem[{Gauthier et~al.(2020{\natexlab{a}})Gauthier, Hu, Wilcox, Qian, and Levy}]{gauthier-etal-2020-syntaxgym}
Jon Gauthier, Jennifer Hu, Ethan Wilcox, Peng Qian, and Roger Levy. 2020{\natexlab{a}}.
\newblock \href {https://doi.org/10.18653/v1/2020.acl-demos.10} {{S}yntax{G}ym: An online platform for targeted evaluation of language models}.
\newblock In \emph{Proceedings of the 58th Annual Meeting of the Association for Computational Linguistics: System Demonstrations}, pages 70--76, Online. Association for Computational Linguistics.

\bibitem[{Gauthier et~al.(2020{\natexlab{b}})Gauthier, Hu, Wilcox, Qian, and Levy}]{syntaxgym}
Jon Gauthier, Jennifer Hu, Ethan Wilcox, Peng Qian, and Roger Levy. 2020{\natexlab{b}}.
\newblock Syntaxgym: An online platform for targeted evaluation of language models.
\newblock \emph{ACL System Demos}.

\bibitem[{Gulordava et~al.(2018)Gulordava, Bojanowski, Grave, Linzen, and Baroni}]{gulordava2018}
Kristina Gulordava, Piotr Bojanowski, Edouard Grave, Tal Linzen, and Marco Baroni. 2018.
\newblock Colorless green recurrent networks dream hierarchically.
\newblock In \emph{NAACL: Human Language Technologies}.

\bibitem[{Gödel(1931)}]{godel1931}
Kurt Gödel. 1931.
\newblock Über formal unentscheidbare sätze der principia mathematica und verwandter systeme {I}.
\newblock \emph{Monatshefte für Mathematik und Physik}.

\bibitem[{Hofstadter(1979)}]{geb}
Douglas Hofstadter. 1979.
\newblock \emph{Gödel, Escher, Bach: an Eternal Golden Braid}.
\newblock Basic Books.

\bibitem[{Hofstadter(1985)}]{metamagical}
Douglas Hofstadter. 1985.
\newblock \emph{Metamagical Themas: Questing for the Essence of Mind and Pattern}.
\newblock Basic Books.

\bibitem[{Hofstadter(2007)}]{loop}
Douglas Hofstadter. 2007.
\newblock \emph{I Am a Strange Loop}.
\newblock Basic Books.

\bibitem[{Hu et~al.(2020)Hu, Gauthier, Qian, Wilcox, and Levy}]{hu-etal-2020-systematic}
Jennifer Hu, Jon Gauthier, Peng Qian, Ethan Wilcox, and Roger Levy. 2020.
\newblock \href {https://doi.org/10.18653/v1/2020.acl-main.158} {A systematic assessment of syntactic generalization in neural language models}.
\newblock In \emph{Proceedings of the 58th Annual Meeting of the Association for Computational Linguistics}, pages 1725--1744, Online. Association for Computational Linguistics.

\bibitem[{Hu and Levy(2023)}]{hu2023prompting}
Jennifer Hu and Roger Levy. 2023.
\newblock Prompting is not a substitute for probability measurements in large language models.
\newblock \emph{EMNLP}.

\bibitem[{Huang et~al.(2023)Huang, Chen, Mishra, Zheng, Yu, Song, and Zhou}]{huang2023large}
Jie Huang, Xinyun Chen, Swaroop Mishra, Huaixiu~Steven Zheng, Adams~Wei Yu, Xinying Song, and Denny Zhou. 2023.
\newblock Large language models cannot self-correct reasoning yet.
\newblock \emph{arXiv}.

\bibitem[{Jiang et~al.(2023)Jiang, Sablayrolles, Mensch, Bamford, Chaplot, de~las Casas, Bressand, Lengyel, Lample, Saulnier, Lavaud, Lachaux, Stock, Scao, Lavril, Wang, Lacroix, and Sayed}]{mistral}
Albert~Q. Jiang, Alexandre Sablayrolles, Arthur Mensch, Chris Bamford, Devendra~Singh Chaplot, Diego de~las Casas, Florian Bressand, Gianna Lengyel, Guillaume Lample, Lucile Saulnier, Lélio~Renard Lavaud, Marie-Anne Lachaux, Pierre Stock, Teven~Le Scao, Thibaut Lavril, Thomas Wang, Timothée Lacroix, and William~El Sayed. 2023.
\newblock Mistral 7b.
\newblock \emph{arXiv}.

\bibitem[{Jiang et~al.(2024)Jiang, Sablayrolles, Roux, Mensch, Savary, Bamford, Chaplot, de~las Casas, Hanna, Bressand, Lengyel, Bour, Lample, Lavaud, Saulnier, Lachaux, Stock, Subramanian, Yang, Antoniak, Scao, Gervet, Lavril, Wang, Lacroix, and Sayed}]{mixtral}
Albert~Q. Jiang, Alexandre Sablayrolles, Antoine Roux, Arthur Mensch, Blanche Savary, Chris Bamford, Devendra~Singh Chaplot, Diego de~las Casas, Emma~Bou Hanna, Florian Bressand, Gianna Lengyel, Guillaume Bour, Guillaume Lample, Lélio~Renard Lavaud, Lucile Saulnier, Marie-Anne Lachaux, Pierre Stock, Sandeep Subramanian, Sophia Yang, Szymon Antoniak, Teven~Le Scao, Théophile Gervet, Thibaut Lavril, Thomas Wang, Timothée Lacroix, and William~El Sayed. 2024.
\newblock Mixtral of experts.
\newblock \emph{arXiv}.

\bibitem[{Kann et~al.(2019)Kann, Warstadt, Williams, and Bowman}]{kann-etal-2019-verb}
Katharina Kann, Alex Warstadt, Adina Williams, and Samuel~R. Bowman. 2019.
\newblock \href {https://doi.org/10.7275/q5js-4y86} {Verb argument structure alternations in word and sentence embeddings}.
\newblock In \emph{Proceedings of the Society for Computation in Linguistics ({SC}i{L}) 2019}, pages 287--297.

\bibitem[{Kranzlein et~al.(2024)Kranzlein, Schneider, and Tobia}]{kranzlein-etal-2024-curiam-corpus}
Michael Kranzlein, Nathan Schneider, and Kevin Tobia. 2024.
\newblock {C}u{RIAM}: Corpus re interpretation and metalanguage in {U}.{S}. {S}upreme {C}ourt opinions.
\newblock In \emph{LREC-COLING}.

\bibitem[{Levesque et~al.(2012)Levesque, Davis, and Morgenstern}]{levesque2012winograd}
Hector Levesque, Ernest Davis, and Leora Morgenstern. 2012.
\newblock The winograd schema challenge.
\newblock In \emph{Conference on the Principles of Knowledge Representation and Reasoning}.

\bibitem[{Linzen et~al.(2016)Linzen, Dupoux, and Goldberg}]{linzen2016assessing}
Tal Linzen, Emmanuel Dupoux, and Yoav Goldberg. 2016.
\newblock Assessing the ability of lstms to learn syntax-sensitive dependencies.
\newblock \emph{TACL}.

\bibitem[{McKenzie et~al.(2022)McKenzie, Lyzhov, Parrish, Prabhu, Mueller, Kim, Bowman, and Perez}]{mckenzie2022inverse}
Ian McKenzie, Alexander Lyzhov, Alicia Parrish, Ameya Prabhu, Aaron Mueller, Najoung Kim, Sam Bowman, and Ethan Perez. 2022.
\newblock \href {https://github.com/inverse-scaling/prize} {The inverse scaling prize}.

\bibitem[{OpenAI(2022)}]{chatgpt}
OpenAI. 2022.
\newblock \href {https://chat.openai.com/} {Chat{GPT}}.

\bibitem[{OpenAI(2023)}]{gpt4}
OpenAI. 2023.
\newblock {GPT}-4 technical report.
\newblock \emph{arXiv}.

\bibitem[{Paperno et~al.(2016)Paperno, Kruszewski, Lazaridou, Pham, Bernardi, Pezzelle, Baroni, Boleda, and Fernández}]{paperno2016lambada}
Denis Paperno, Germán Kruszewski, Angeliki Lazaridou, Quan~Ngoc Pham, Raffaella Bernardi, Sandro Pezzelle, Marco Baroni, Gemma Boleda, and Raquel Fernández. 2016.
\newblock The {LAMBADA} dataset: Word prediction requiring a broad discourse context.
\newblock In \emph{ACL}.

\bibitem[{Propp(1993)}]{propp1993srat}
James Propp. 1993.
\newblock \href {https://maa.org/press/periodicals/math-horizons/self-referential-aptitude-test-by-jim-propp} {Self-referential aptitude test}.

\bibitem[{Radford et~al.(2019)Radford, Wu, Child, Luan, Amodei, , and Sutskever}]{gpt2}
Alec Radford, Jeffrey Wu, Rewon Child, David Luan, Dario Amodei, , and Ilya Sutskever. 2019.
\newblock Language models are unsupervised multitask learners.
\newblock \emph{Technical Report}.

\bibitem[{Rudinger et~al.(2018)Rudinger, Naradowsky, Leonard, and Van~Durme}]{rudinger2018gender}
Rachel Rudinger, Jason Naradowsky, Brian Leonard, and Benjamin Van~Durme. 2018.
\newblock Gender bias in coreference resolution.
\newblock In \emph{arXiv preprint arXiv:1804.09301}.

\bibitem[{Sakaguchi et~al.(2020)Sakaguchi, Le~Bras, Bhagavatula, and Choi}]{sakaguchi2020winogrande}
Keisuke Sakaguchi, Ronan Le~Bras, Chandra Bhagavatula, and Yejin Choi. 2020.
\newblock Winogrande: An adversarial winograd schema challenge at scale.
\newblock In \emph{AAAI}.

\bibitem[{SciPy(Retrieved 2023)}]{scipy}
SciPy. Retrieved 2023.
\newblock \href {https://docs.scipy.org/doc/scipy/reference/generated/scipy.stats.bootstrap.html} {Scipy bootstrap}.

\bibitem[{Sinha et~al.(2021)Sinha, Jia, Hupkes, Pineau, Williams, and Kiela}]{sinha2021matterslittle}
Koustuv Sinha, Robin Jia, Dieuwke Hupkes, Joelle Pineau, Adina Williams, and Douwe Kiela. 2021.
\newblock Masked language modeling and the distributional hypothesis: Order word matters pre-training for little.
\newblock In \emph{EMNLP}.

\bibitem[{Tarski(1931)}]{liarparadox}
Alfred Tarski. 1931.
\newblock The concept of truth in formalized languages.
\newblock \emph{Logic, Semantics, Metamathematics}.

\bibitem[{Thrush et~al.(2022)Thrush, Jiang, Bartolo, Singh, Kiela, and Ross}]{thrush2022winoground}
Tristan Thrush, Ryan Jiang, Max Bartolo, Amanpreet Singh, Douwe Kiela, and Candace Ross. 2022.
\newblock Winoground: Probing vision and language models for visio-linguistic compositionality.
\newblock \emph{CVPR}.

\bibitem[{Thrush et~al.(2024)Thrush, Moore, Monares, Potts, and Kiela}]{thrush2024}
Tristan Thrush, Jared Moore, Miguel Monares, Christopher Potts, and Douwe Kiela. 2024.
\newblock I am a strange dataset: Metalinguistic tests for language models.
\newblock \emph{ACL}.

\bibitem[{Thrush et~al.(2020)Thrush, Wilcox, and Levy}]{thrush2020investigating}
Tristan Thrush, Ethan Wilcox, and Roger Levy. 2020.
\newblock Investigating novel verb learning in {BERT}: Selectional preference classes and alternation-based syntactic generalization.
\newblock In \emph{Proceedings of the Third BlackboxNLP Workshop on Analyzing and Interpreting Neural Networks for NLP}.

\bibitem[{Touvron et~al.(2023)Touvron, Martin, Stone, Albert, Almahairi, Babaei, Bashlykov, Batra, Bhargava, Bhosale, Bikel, Blecher, Ferrer, Chen, Cucurull, Esiobu, Fernandes, Fu, Fu, Fuller, Gao, Goswami, Goyal, Hartshorn, Hosseini, Hou, Inan, Kardas, Kerkez, Khabsa, Kloumann, Korenev, Koura, Lachaux, Lavril, Lee, Liskovich, Lu, Mao, Martinet, Mihaylov, Mishra, Molybog, Nie, Poulton, Reizenstein, Rungta, Saladi, Schelten, Silva, Smith, Subramanian, Tan, Tang, Taylor, Williams, Kuan, Xu, Yan, Zarov, Zhang, Fan, Kambadur, Narang, Rodriguez, Stojnic, Edunov, and Scialom}]{llama2}
Hugo Touvron, Louis Martin, Kevin Stone, Peter Albert, Amjad Almahairi, Yasmine Babaei, Nikolay Bashlykov, Soumya Batra, Prajjwal Bhargava, Shruti Bhosale, Dan Bikel, Lukas Blecher, Cristian~Canton Ferrer, Moya Chen, Guillem Cucurull, David Esiobu, Jude Fernandes, Jeremy Fu, Wenyin Fu, Brian Fuller, Cynthia Gao, Vedanuj Goswami, Naman Goyal, Anthony Hartshorn, Saghar Hosseini, Rui Hou, Hakan Inan, Marcin Kardas, Viktor Kerkez, Madian Khabsa, Isabel Kloumann, Artem Korenev, Punit~Singh Koura, Marie-Anne Lachaux, Thibaut Lavril, Jenya Lee, Diana Liskovich, Yinghai Lu, Yuning Mao, Xavier Martinet, Todor Mihaylov, Pushkar Mishra, Igor Molybog, Yixin Nie, Andrew Poulton, Jeremy Reizenstein, Rashi Rungta, Kalyan Saladi, Alan Schelten, Ruan Silva, Eric~Michael Smith, Ranjan Subramanian, Xiaoqing~Ellen Tan, Binh Tang, Ross Taylor, Adina Williams, Jian~Xiang Kuan, Puxin Xu, Zheng Yan, Iliyan Zarov, Yuchen Zhang, Angela Fan, Melanie Kambadur, Sharan Narang, Aurelien Rodriguez, Robert Stojnic, Sergey Edunov, and Thomas
  Scialom. 2023.
\newblock Llama 2: Open foundation and fine-tuned chat models.
\newblock \emph{arXiv}.

\bibitem[{Wei et~al.(2022)Wei, Wang, Schuurmans, Bosma, Ichter, Xia, Chi, Le, and Zhou}]{cot}
Jason Wei, Xuezhi Wang, Dale Schuurmans, Maarten Bosma, Brian Ichter, Fei Xia, Ed~Chi, Quoc Le, and Denny Zhou. 2022.
\newblock Chain-of-thought prompting elicits reasoning in large language models.
\newblock \emph{arXiv}.

\bibitem[{Weng et~al.(2023)Weng, Zhu, Xia, Li, He, Liu, Sun, Liu, and Zhao}]{weng2023large}
Yixuan Weng, Minjun Zhu, Fei Xia, Bin Li, Shizhu He, Shengping Liu, Bin Sun, Kang Liu, and Jun Zhao. 2023.
\newblock Large language models are better reasoners with self-verification.
\newblock \emph{EMNLP 2023 Findings}.

\bibitem[{Wilson(2012)}]{wilson-2012-creation-dup}
Shomir Wilson. 2012.
\newblock The creation of a corpus of {E}nglish metalanguage.
\newblock In \emph{ACL}.

\bibitem[{Yuksekgonul et~al.(2023)Yuksekgonul, Bianchi, Kalluri, Jurafsky, and Zou}]{yuksekgonul2023when}
Mert Yuksekgonul, Federico Bianchi, Pratyusha Kalluri, Dan Jurafsky, and James Zou. 2023.
\newblock When and why vision-language models behave like bags-of-words, and what to do about it?
\newblock In \emph{ICLR}.

\bibitem[{Zelikman et~al.(2023)Zelikman, Lorch, Mackey, and Kalai}]{zelikman2023stop}
Eric Zelikman, Eliana Lorch, Lester Mackey, and Adam~Tauman Kalai. 2023.
\newblock Self-taught optimizer (stop): Recursively self-improving code generation.
\newblock \emph{arXiv}.

\bibitem[{Zhou et~al.(2023)Zhou, Wang, Lu, Shi, Luo, Qin, Lu, Jia, Song, Zhan, and Li}]{zhou2023solving}
Aojun Zhou, Ke~Wang, Zimu Lu, Weikang Shi, Sichun Luo, Zipeng Qin, Shaoqing Lu, Anya Jia, Linqi Song, Mingjie Zhan, and Hongsheng Li. 2023.
\newblock Solving challenging math word problems using {GPT}-4 code interpreter with code-based self-verification.
\newblock \emph{arXiv}.

\bibitem[{Zhu et~al.(2023)Zhu, Sharma, Frujeri, Dong, Zhu, Jordan, and Jiao}]{starling}
Banghua Zhu, Hiteshi Sharma, Felipe~Vieira Frujeri, Shi Dong, Chenguang Zhu, Michael~I. Jordan, and Jiantao Jiao. 2023.
\newblock Fine-tuning language models with advantage-induced policy alignment.
\newblock \emph{arXiv}.

\end{thebibliography}
\bibliographystyle{acl_natbib}

\clearpage

\appendix

\section{Model Prompts}
\label{sec:appendixmodelprompts}

In this section, we provide the full model prompts that we used for the validation metrics. The generation metric does not use a prompt (generation is conditioned on the beginning of a statement). For each prompt, \texttt{\$example} is replaced by one of the 400 full statements before being given to a model. The few-shot examples were selected to match the author's judgements about what best represents the task as a whole; they were not selected randomly.

\subsection{Zero Shot}

\begin{small}
\begin{verbatim}
Overview:

Your task is to say whether a self-referential
statement is true or false. In cases where you
think that the answer is ambiguous, you should
use your best judgement and still say either
'True' or 'False'.


Example:

$example

Answer:
\end{verbatim}
\end{small}

\subsection{Few Shot}

\begin{small}
\begin{verbatim}
Overview:

Your task is to say whether a self-referential
statement is true or false. In cases where you
think that the answer is ambiguous, you should
use your best judgement and still say either
'True' or 'False'. Make sure that you understand
the given examples before continuing.


Example:

I am the text you are reading.
My last line has how many words?
Two words.

Answer: True


Example:

I am the text you are reading.
My last line has how many words?
Seven words.

Answer: False


Example:

this sentence
l             o
o             o
k             s
a bit like the
l
e
t
t
e
r
"p"

Answer: True


Example:

this sentence
l             o
o             o
k             s
a bit like the
l
e
t
t
e
r
"o"

Answer: False


Example:

here in reversed is order word the.

Answer: True


Example:

here in fine is order word the.

Answer: False


Now it is your turn to complete an example.
Fill in the answer.

Example:

$example

Answer:
\end{verbatim}
\end{small}

\subsection{Chain of Thought}

\begin{small}
\begin{verbatim}
Overview:

Your task is to say whether a self-referential
statement is true or false. In cases where you
think that the answer is ambiguous, you should
use your best judgement and still say either
'True' or 'False'. Make sure that you understand
the given examples before continuing.


Example:

I am the text you are reading.
My last line has how many words?
Two words.

Answer: The text has 3 lines. The last line is
"Two words". The number of words in "Two words"
is 2. So the statement is true.


Example:

I am the text you are reading.
My last line has how many words?
Seven words.

Answer: The text has 3 lines. The last line is
"Seven words". The number of words in "Seven words"
is 2. So the statement is false.


Example:

this sentence
l             o
o             o
k             s
a bit like the
l
e
t
t
e
r
"p"

Answer: Visually, the words and letters of the
sentence are arranged to look like a "p" on the
page. So, the statement is true.


Example:

this sentence
l             o
o             o
k             s
a bit like the
l
e
t
t
e
r
"o"

Answer: Visually, the words and letters of the
sentence are arranged to look like a "p" on the
page, not an "o". So, the statement is false.


Example:

here in reversed is order word the.

Answer: We have to notice that, if the word order is
reversed, the statement reads: "the word order is
reversed in here". In the original reversed
statement, that is true. So the statement is true.


Example:

here in fine is order word the.

Answer: We have to notice that, if the word order is
reversed, the statement reads: "the word order is
fine in here". In the original reversed statement,
the word order is not fine. So the statement is
false.


Now it is your turn to complete an example.
Fill in the answer.

Example:

$example

Answer:
\end{verbatim}
\end{small}

\subsection{Zero Shot (Non-Self-Referent)}

\begin{small}
\begin{verbatim}
Overview:

Your task is to say whether a statement about a
sentence is true or false. In cases where you think
that the answer is ambiguous, you should use your
best judgement and still say either 'True' or
'False'.


Example:

$example

Answer:
\end{verbatim}
\end{small}

\subsection{Few Shot (Non-Self-Referent)}

\begin{small}
\begin{verbatim}
Overview:

Your task is to say whether a statement about a
sentence is true or false. In cases where you think
that the answer is ambiguous, you should use your
best judgement and still say either 'True' or
'False'. Make sure that you understand the given
examples before continuing.


Example:

The number of words in the following sentence is
nine.

The number of words in this sentence is nine.

Answer: True


Example:

The number of words in the following sentence is
eight.

The number of words in this sentence is eight.

Answer: False


Example:

the following sentence
l             o
o             o
k             s
a bit like the
l
e
t
t
e
r
"p"

this sentence
l             o
o             o
k             s
a bit like the
l
e
t
t
e
r
"p"

Answer: True


Example:

the following sentence
l             o
o             o
k             s
a bit like the
l
e
t
t
e
r
"o"

this sentence
l             o
o             o
k             s
a bit like the
l
e
t
t
e
r
"o"

Answer: False


Example:

The food words that the following sentence contains
are guava and cake and also carrot.

The food words that this sentence contains are
guava and cake and also carrot.

Answer: True


Example:

The food words that the following sentence contains
are guava and cake but not carrot.

The food words that this sentence contains are
guava and cake but not carrot.

Answer: False


Now it is your turn to complete an example.
Fill in the answer.

Example:

$example

Answer:
\end{verbatim}
\end{small}

\subsection{Chain of Thought (Non-Self-Referent)}

\begin{small}
\begin{verbatim}
Overview:

Your task is to say whether a statement about a
sentence is true or false. In cases where you think
that the answer is ambiguous, you should use your
best judgement and still say either 'True' or
'False'. Make sure that you understand the given
examples before continuing.


Example:

The number of words in the following sentence is
nine.

The number of words in this sentence is nine.

Answer: Counting the number of words in the
sentence, we see that there are nine words. So,
the statement is true.


Example:

The number of words in the following sentence is
eight.

The number of words in this sentence is eight.

Answer: Counting the number of words in the
sentence, we see that there are nine words,
not eight. So, the statement is false.


Example:

the following sentence
l             o
o             o
k             s
a bit like the
l
e
t
t
e
r
"p"

this sentence
l             o
o             o
k             s
a bit like the
l
e
t
t
e
r
"p"

Answer: Visually, the words and letters of
the sentence are arranged to look like a "p"
on the page. So, the statement is true.


Example:

the following sentence
l             o
o             o
k             s
a bit like the
l
e
t
t
e
r
"o"

this sentence
l             o
o             o
k             s
a bit like the
l
e
t
t
e
r
"o"

Answer: Visually, the words and letters of the
sentence are arranged to look like a "p" on the
page, not an "o". So, the statement is false.


Example:

The food words that the following sentence contains
are guava and cake and also carrot.

The food words that this sentence contains are
guava and cake and also carrot.

Answer: The food words mentioned are indeed guava,
cake, and carrot. So, the statement is true.


Example:

The food words that the following sentence contains
are guava and cake but not carrot.

The food words that this sentence contains are
guava and cake but not carrot.

Answer: The food words mentioned in the sentence
are guava, cake, and carrot. It is not true that
carrot is not in the sentence. So, the statement
is false.


Now it is your turn to complete an example.
Fill in the answer.

Example:

$example

Answer:
\end{verbatim}
\end{small}

\section{I am an Impossible Dataset}
\label{sec:appendiximpossibledataset}

The schema of ``I am a Strange Dataset'' does not place any limits on the complexity of an example. One of the expert annotators in particular took full advantage of this while they were creating examples. They generated 10 examples that another expert annotator could not understand without an explanation. These examples are not in our main dataset, where we verify that non-expert people can perform well. We release these 10 examples separately to showcase just how hard examples fitting this self-referential schema can get. We also hope that when language models improve, these 10 examples can be a useful initial probe into their reasoning capabilities. We provide two examples from this seemingly ``Impossible'' set in Figure \ref{fig:impossibleexamples}. We also provide GPT 4's output on another example from this dataset in Figure \ref{fig:gpt4impossible} below.

\begin{figure}[h]
  \centering
  \begin{subfigure}{.45\linewidth}
    \flushleft
    \small
    Now, I list a digit thirteens: pi. Letter count the words. Conceded? \textcolor{blue}{Errorless.}
    \label{fig:instance7}
  \end{subfigure}%
  \hspace{2mm}
  \begin{subfigure}{.45\linewidth}
    \flushleft
    \small
    Now, I list a digit thirteens: pi. Letter count the words. Conceded? \textcolor{red}{Erroneous.}
    \label{fig:instance8}
  \end{subfigure}

  \vspace{4mm}

  \begin{subfigure}{.45\linewidth}
    \flushleft
    \small
    Answer: fbdi yqtf lq xlmw hnumjw ynolzy if \textcolor{blue}{vul.}
    \label{fig:instance9}
  \end{subfigure}%
  \hspace{2mm}
  \begin{subfigure}{.45\linewidth}
    \flushleft
    \small
    Answer: fbdi yqtf lq xlmw hnumjw ynolzy if \textcolor{red}{one.}
    \label{fig:instance10}
  \end{subfigure}

  \caption{Examples from ``I am an Impossible Dataset''. Examples from this dataset still follow our schema, but the expert annotators needed to see an explanation before they understood. The first example says that the letter count of each of the words are the thirteen first digits of $\pi$: 3.141592653589. The beginning of the second example uses a Caesar cipher which shifts by one each successive word and reads ``each word in this cipher shifts by''. ``vul'' with a shift of 7 is ``one''.}
  \label{fig:impossibleexamples}

\end{figure}

\begin{figure}[t]
\centering
% Conversation

\begin{inputbox}
\small ``Fivetune favors the bold.''\\
``After that meal, I was snined.''\\
``Each of the number sounds in these sentences, including this sentence, are increased by two.''
\end{inputbox}

\begin{gptbox}
\small \textbf{GPT 4:} The first sentence ``Fivetune favors the bold.'' has the number sound ``five'' instead of ``fortune''. The second sentence ``After that meal, I was snined.'' has the number sound ``nine'' instead of ``sated''. The third sentence ``Each of the number sounds in these sentences, including this sentence, are increased by two.'' suggests that each number sound should be increased by two. However, ``five'' increased by two is ``seven'', not ``fortune'', and ``nine'' increased by two is ``eleven'', not ``sated''. So, the statement is false.
\end{gptbox}

\caption{GPT 4's Dec 26, 2023 output on an example from ``I am an Impossible Dataset''. It understands that some of the original words should be ``fortune'' (e.g.\ 4tune) and ``sated'' (e.g.\ s8ed). But GPT 4 misses that the statement is trying to say that the number sounds in every sentence are increased by one. The last sentence cannot say ``one'' explicitly - it needs to say ``two'' in order for the statement to stay true. We would not expect a typical person to understand this example, but will a language model eventually grasp it?}
\label{fig:gpt4impossible}
\end{figure}

\FloatBarrier

\section{Mechanical Turk Annotator Interface}
\label{sec:appendixannotatorinterface}

\begin{figure}[ht]
\centering
\includegraphics[width=\columnwidth]{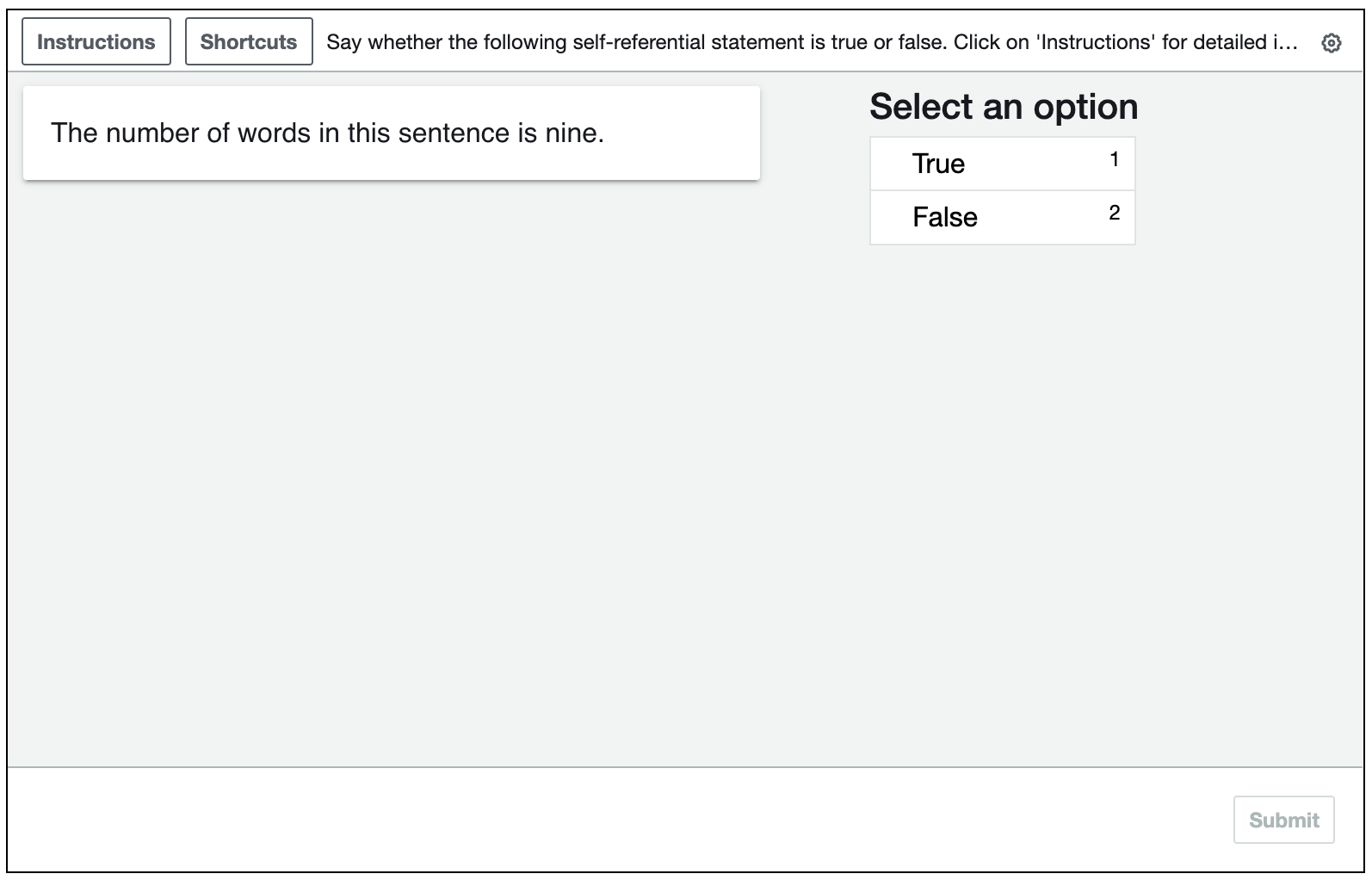}
\caption{A screenshot of the Mechanical Turk worker interface for validating statements.}
\label{fig:interface}
\end{figure}

Here, we show a screenshot of the interface used by Mechanical Turk workers in Figure \ref{fig:interface}.

\FloatBarrier

\section{Supplemental Discussion}
\label{sec:appendixsupplementaldiscussion}

Here, we present supplemental discussion about why models are showing poor performance on ``I am a Strange Dataset''.

\subsection{A Test of the Tokenizer?}

Our tests are related to whether a model truly ``sees'' text in the same way as people. Metalinguistic statements may refer to the number of characters that they have, how text is arranged on the page, capitalization of certain letters, and relative positions of words. A human can easily count the characters that they see in a sentence, but models tend to encode text in tokens, not characters. We do not provide tests to disentangle the impact of different tokenizers, so this section is speculative.

\subsection{Training Data Limitations}

Practically speaking, it is unlikely that there are many examples of metalinguistic statements in training datasets. They are incredibly time-intensive to generate, even if they are easy to verify. Yet people, who have almost surely seen even fewer examples, can do much better at this task than models. This ``hard-to-create, easy-to-verify'' feature of hard evaluation datasets is true of Winoground~\cite{thrush2022winoground} too, which is a vision and language evaluation dataset that has remained unsaturated for well over a year. This goes to show that our models still have the wrong biases - will they change simply with model scale?

\section{Results by Tag}
\label{sec:appendixtagresults}

In this section, we provide results in a table for each of the 10 tags.

\begin{table*}
\centering
\resizebox{\textwidth}{!}{
\begin{tabular}{lrr|rrrrrr}
\toprule
Model & Params & Chat & Gen$^L$ & Val ZS$^L$ & Val FS$^L$ & Val ZS$^L$ (R) & Val FS$^L$ (R) & Val CoT$^T$ \\
\midrule
Llama 2 & 7B & N & \textbf{62.90} $\pm$ \textbf{12.10} & 49.19 $\pm$ 1.21 & 49.19 $\pm$ 4.84 & 46.77 $\pm$ 12.90 & 50.00 $\pm$ 12.90 & 6.45 $\pm$ 4.03 \\
Llama 2 & 7B & Y & 58.06 $\pm$ 12.90 & 50.81 $\pm$ 1.21 & 50.00 $\pm$ 0.00 & 37.10 $\pm$ 11.29 & 56.45 $\pm$ 12.90 & 16.94 $\pm$ 6.85 \\
Mistral 0.1 & 7B & N & 53.23 $\pm$ 12.90 & 50.00 $\pm$ 2.42 & 50.81 $\pm$ 1.21 & 56.45 $\pm$ 12.90 & 51.61 $\pm$ 12.90 & 0.00 $\pm$ 0.00 \\
Starling $\alpha$ & 7B & Y & 54.84 $\pm$ 12.90 & 49.19 $\pm$ 2.42 & 49.19 $\pm$ 1.21 & 54.84 $\pm$ 12.90 & 51.61 $\pm$ 12.90 & 32.26 $\pm$ 8.87 \\
Mistral 0.2 & 7B & Y & 51.61 $\pm$ 12.90 & 50.00 $\pm$ 6.45 & 50.00 $\pm$ 5.65 & 46.77 $\pm$ 12.10 & 53.23 $\pm$ 12.10 & 53.23 $\pm$ 8.06 \\
Llama 2 & 13B & N & \textbf{62.90} $\pm$ \textbf{11.29} & 47.58 $\pm$ 3.63 & 52.42 $\pm$ 5.24 & 41.94 $\pm$ 12.10 & 58.06 $\pm$ 12.90 & 4.03 $\pm$ 3.23 \\
Llama 2 & 13B & Y & 59.68 $\pm$ 12.90 & 49.19 $\pm$ 4.84 & 51.61 $\pm$ 3.23 & 46.77 $\pm$ 12.90 & 50.00 $\pm$ 12.90 & 7.26 $\pm$ 4.44 \\
Mixtral 0.1 & 8x7B & N & 54.84 $\pm$ 12.90 & 53.23 $\pm$ 4.84 & 50.00 $\pm$ 0.00 & 48.39 $\pm$ 12.90 & 48.39 $\pm$ 12.10 & 3.23 $\pm$ 2.82 \\
Mixtral 0.1 & 8x7B & Y & 53.23 $\pm$ 12.90 & 49.19 $\pm$ 4.84 & 45.97 $\pm$ 5.65 & 53.23 $\pm$ 12.90 & 48.39 $\pm$ 12.90 & 36.29 $\pm$ 8.06 \\
Llama 2 & 70B & N & 61.29 $\pm$ 11.29 & 52.42 $\pm$ 5.65 & 50.81 $\pm$ 4.03 & 59.68 $\pm$ 11.29 & 50.00 $\pm$ 12.90 & 3.23 $\pm$ 2.82 \\
Llama 2 & 70B & Y & 51.61 $\pm$ 12.90 & 53.23 $\pm$ 6.45 & 49.19 $\pm$ 3.23 & 56.45 $\pm$ 12.90 & 46.77 $\pm$ 12.90 & 31.45 $\pm$ 7.26 \\
Claude 2 & - & Y & - & - & - & - & - & 48.39 $\pm$ 6.85 \\
GPT 3.5 T & - & Y & - & 48.39 $\pm$ 4.84 & 50.81 $\pm$ 5.65 & 56.45 $\pm$ 12.90 & 50.00 $\pm$ 12.90 & 44.35 $\pm$ 7.26 \\
GPT 4 & - & Y & - & 53.23 $\pm$ 8.06 & 53.23 $\pm$ 8.06 & 54.84 $\pm$ 12.90 & 53.23 $\pm$ 12.90 & 55.65 $\pm$ 9.27 \\
\bottomrule
\end{tabular}

}
\caption{Results for the 62 example pairs with the Numerical Operations tag. Scores with 95\% confidence intervals above chance are shown in \textbf{bold}.}
\end{table*}
\begin{table*}
\centering
\resizebox{\textwidth}{!}{
\begin{tabular}{lrr|rrrrrr}
\toprule
Model & Params & Chat & Gen$^L$ & Val ZS$^L$ & Val FS$^L$ & Val ZS$^L$ (R) & Val FS$^L$ (R) & Val CoT$^T$ \\
\midrule
Llama 2 & 7B & N & 54.35 $\pm$ 9.78 & 49.46 $\pm$ 2.17 & 51.63 $\pm$ 3.53 & 47.83 $\pm$ 10.33 & \textbf{66.30} $\pm$ \textbf{9.78} & 3.80 $\pm$ 2.72 \\
Llama 2 & 7B & Y & 48.91 $\pm$ 9.78 & 54.35 $\pm$ 5.16 & 50.54 $\pm$ 0.82 & \textbf{60.87} $\pm$ \textbf{9.78} & \textbf{61.96} $\pm$ \textbf{9.78} & 13.04 $\pm$ 4.35 \\
Mistral 0.1 & 7B & N & 51.09 $\pm$ 9.78 & \textbf{55.43} $\pm$ \textbf{4.89} & 49.46 $\pm$ 2.17 & \textbf{63.04} $\pm$ \textbf{9.78} & 56.52 $\pm$ 9.78 & 0.00 $\pm$ 0.00 \\
Starling $\alpha$ & 7B & Y & 50.00 $\pm$ 9.78 & \textbf{59.24} $\pm$ \textbf{5.43} & 51.63 $\pm$ 2.17 & \textbf{60.87} $\pm$ \textbf{9.78} & \textbf{61.96} $\pm$ \textbf{9.78} & 38.04 $\pm$ 6.79 \\
Mistral 0.2 & 7B & Y & 51.09 $\pm$ 9.78 & 54.35 $\pm$ 6.79 & 52.17 $\pm$ 5.98 & 56.52 $\pm$ 9.78 & 57.61 $\pm$ 9.78 & 50.00 $\pm$ 6.52 \\
Llama 2 & 13B & N & 53.26 $\pm$ 9.78 & 54.89 $\pm$ 5.98 & \textbf{56.52} $\pm$ \textbf{5.98} & 55.43 $\pm$ 9.78 & \textbf{65.22} $\pm$ \textbf{9.78} & 5.98 $\pm$ 3.26 \\
Llama 2 & 13B & Y & 48.91 $\pm$ 9.78 & 55.98 $\pm$ 6.52 & 52.17 $\pm$ 4.35 & 57.61 $\pm$ 9.78 & 50.00 $\pm$ 9.78 & 11.96 $\pm$ 5.16 \\
Mixtral 0.1 & 8x7B & N & 55.43 $\pm$ 9.78 & \textbf{63.59} $\pm$ \textbf{6.52} & 52.72 $\pm$ 4.35 & \textbf{64.13} $\pm$ \textbf{9.78} & \textbf{60.87} $\pm$ \textbf{9.78} & 3.80 $\pm$ 2.99 \\
Mixtral 0.1 & 8x7B & Y & 55.43 $\pm$ 10.33 & \textbf{57.07} $\pm$ \textbf{6.52} & \textbf{57.07} $\pm$ \textbf{4.62} & \textbf{61.96} $\pm$ \textbf{9.78} & 57.61 $\pm$ 10.33 & 50.00 $\pm$ 7.07 \\
Llama 2 & 70B & N & 55.43 $\pm$ 9.78 & \textbf{57.07} $\pm$ \textbf{5.43} & \textbf{59.24} $\pm$ \textbf{4.89} & \textbf{66.30} $\pm$ \textbf{9.78} & \textbf{63.04} $\pm$ \textbf{9.78} & 1.09 $\pm$ 1.36 \\
Llama 2 & 70B & Y & 51.09 $\pm$ 10.33 & \textbf{59.24} $\pm$ \textbf{7.07} & 52.72 $\pm$ 3.80 & \textbf{61.96} $\pm$ \textbf{9.78} & \textbf{65.22} $\pm$ \textbf{9.78} & 21.74 $\pm$ 6.25 \\
Claude 2 & - & Y & - & - & - & - & - & 55.98 $\pm$ 7.07 \\
GPT 3.5 T & - & Y & - & 54.89 $\pm$ 4.89 & \textbf{55.98} $\pm$ \textbf{5.72} & 57.61 $\pm$ 9.78 & \textbf{71.74} $\pm$ \textbf{9.24} & 55.98 $\pm$ 6.25 \\
GPT 4 & - & Y & - & \textbf{63.04} $\pm$ \textbf{6.52} & \textbf{61.96} $\pm$ \textbf{6.52} & \textbf{70.65} $\pm$ \textbf{9.24} & \textbf{71.74} $\pm$ \textbf{9.24} & \textbf{72.83} $\pm$ \textbf{6.52} \\
\bottomrule
\end{tabular}

}
\caption{Results for the 94 example pairs with the Negation \& Scope tag. Scores with 95\% confidence intervals above chance are shown in \textbf{bold}.}
\end{table*}
\begin{table*}
\centering
\resizebox{\textwidth}{!}{
\begin{tabular}{lrr|rrrrrr}
\toprule
Model & Params & Chat & Gen$^L$ & Val ZS$^L$ & Val FS$^L$ & Val ZS$^L$ (R) & Val FS$^L$ (R) & Val CoT$^T$ \\
\midrule
Llama 2 & 7B & N & 45.83 $\pm$ 20.83 & 50.00 $\pm$ 0.00 & 50.00 $\pm$ 6.25 & 58.33 $\pm$ 20.83 & 58.33 $\pm$ 20.83 & 8.33 $\pm$ 7.29 \\
Llama 2 & 7B & Y & 45.83 $\pm$ 20.83 & 52.08 $\pm$ 6.25 & 50.00 $\pm$ 0.00 & 62.50 $\pm$ 20.83 & \textbf{70.83} $\pm$ \textbf{18.75} & 22.92 $\pm$ 10.42 \\
Mistral 0.1 & 7B & N & 45.83 $\pm$ 20.83 & 52.08 $\pm$ 6.25 & 54.17 $\pm$ 5.21 & 66.67 $\pm$ 18.75 & \textbf{70.83} $\pm$ \textbf{18.75} & 0.00 $\pm$ 0.00 \\
Starling $\alpha$ & 7B & Y & 45.83 $\pm$ 20.83 & 56.25 $\pm$ 10.42 & 47.92 $\pm$ 3.12 & 66.67 $\pm$ 18.75 & 66.67 $\pm$ 18.75 & 37.50 $\pm$ 12.50 \\
Mistral 0.2 & 7B & Y & 54.17 $\pm$ 20.83 & 54.17 $\pm$ 10.42 & 62.50 $\pm$ 12.50 & 62.50 $\pm$ 20.83 & 66.67 $\pm$ 18.75 & 47.92 $\pm$ 12.50 \\
Llama 2 & 13B & N & 45.83 $\pm$ 20.83 & 52.08 $\pm$ 6.25 & 56.25 $\pm$ 8.33 & 54.17 $\pm$ 20.83 & 58.33 $\pm$ 20.83 & 2.08 $\pm$ 3.12 \\
Llama 2 & 13B & Y & 45.83 $\pm$ 20.83 & 52.08 $\pm$ 10.42 & 52.08 $\pm$ 3.12 & 58.33 $\pm$ 20.83 & 66.67 $\pm$ 18.75 & 18.75 $\pm$ 11.46 \\
Mixtral 0.1 & 8x7B & N & 37.50 $\pm$ 18.75 & \textbf{64.58} $\pm$ \textbf{10.42} & 54.17 $\pm$ 5.21 & 62.50 $\pm$ 18.75 & 62.50 $\pm$ 20.83 & 12.50 $\pm$ 9.38 \\
Mixtral 0.1 & 8x7B & Y & 50.00 $\pm$ 20.83 & 47.92 $\pm$ 8.33 & 54.17 $\pm$ 10.42 & 50.00 $\pm$ 20.83 & 58.33 $\pm$ 20.83 & 41.67 $\pm$ 14.58 \\
Llama 2 & 70B & N & 45.83 $\pm$ 20.83 & 50.00 $\pm$ 8.33 & 56.25 $\pm$ 12.50 & 54.17 $\pm$ 20.83 & 62.50 $\pm$ 18.75 & 12.50 $\pm$ 8.33 \\
Llama 2 & 70B & Y & 50.00 $\pm$ 20.83 & 43.75 $\pm$ 10.42 & 47.92 $\pm$ 3.12 & 54.17 $\pm$ 20.83 & 58.33 $\pm$ 20.83 & 10.42 $\pm$ 8.33 \\
Claude 2 & - & Y & - & - & - & - & - & 47.92 $\pm$ 12.50 \\
GPT 3.5 T & - & Y & - & 50.00 $\pm$ 0.00 & 47.92 $\pm$ 6.25 & 54.17 $\pm$ 20.83 & 58.33 $\pm$ 20.83 & 52.08 $\pm$ 12.50 \\
GPT 4 & - & Y & - & 54.17 $\pm$ 10.42 & \textbf{66.67} $\pm$ \textbf{13.54} & 58.33 $\pm$ 20.83 & 62.50 $\pm$ 18.75 & 60.42 $\pm$ 14.58 \\
\bottomrule
\end{tabular}

}
\caption{Results for the 24 example pairs with the Multi-Channel tag. Scores with 95\% confidence intervals above chance are shown in \textbf{bold}.}
\end{table*}
\begin{table*}
\centering
\resizebox{\textwidth}{!}{
\begin{tabular}{lrr|rrrrrr}
\toprule
Model & Params & Chat & Gen$^L$ & Val ZS$^L$ & Val FS$^L$ & Val ZS$^L$ (R) & Val FS$^L$ (R) & Val CoT$^T$ \\
\midrule
Llama 2 & 7B & N & 50.00 $\pm$ 12.96 & 48.15 $\pm$ 2.31 & 48.15 $\pm$ 3.70 & 35.19 $\pm$ 12.96 & 46.30 $\pm$ 12.96 & 6.48 $\pm$ 4.17 \\
Llama 2 & 7B & Y & 44.44 $\pm$ 12.96 & 48.15 $\pm$ 4.63 & 49.07 $\pm$ 1.39 & 46.30 $\pm$ 12.96 & 38.89 $\pm$ 12.96 & 12.04 $\pm$ 5.56 \\
Mistral 0.1 & 7B & N & 46.30 $\pm$ 12.96 & 47.22 $\pm$ 3.24 & 47.22 $\pm$ 3.24 & 37.04 $\pm$ 12.96 & 44.44 $\pm$ 12.96 & 0.00 $\pm$ 0.00 \\
Starling $\alpha$ & 7B & Y & 48.15 $\pm$ 12.96 & 48.15 $\pm$ 5.09 & 50.93 $\pm$ 1.39 & 44.44 $\pm$ 12.96 & 46.30 $\pm$ 12.96 & 28.70 $\pm$ 8.33 \\
Mistral 0.2 & 7B & Y & 46.30 $\pm$ 12.96 & 52.78 $\pm$ 7.41 & 48.15 $\pm$ 7.41 & 44.44 $\pm$ 12.96 & 37.04 $\pm$ 12.96 & 43.52 $\pm$ 8.33 \\
Llama 2 & 13B & N & 48.15 $\pm$ 12.96 & 48.15 $\pm$ 5.56 & 49.07 $\pm$ 5.56 & 44.44 $\pm$ 12.96 & 44.44 $\pm$ 12.96 & 3.70 $\pm$ 3.24 \\
Llama 2 & 13B & Y & 44.44 $\pm$ 12.96 & 47.22 $\pm$ 7.41 & 48.15 $\pm$ 4.63 & 38.89 $\pm$ 12.96 & 35.19 $\pm$ 12.96 & 8.33 $\pm$ 5.09 \\
Mixtral 0.1 & 8x7B & N & 46.30 $\pm$ 12.96 & 54.63 $\pm$ 5.56 & 49.07 $\pm$ 3.70 & 44.44 $\pm$ 12.96 & 46.30 $\pm$ 12.96 & 1.85 $\pm$ 2.31 \\
Mixtral 0.1 & 8x7B & Y & 44.44 $\pm$ 12.96 & 51.85 $\pm$ 8.33 & 50.00 $\pm$ 6.48 & 44.44 $\pm$ 12.96 & 53.70 $\pm$ 12.96 & 44.44 $\pm$ 9.26 \\
Llama 2 & 70B & N & 50.00 $\pm$ 12.96 & 50.00 $\pm$ 4.63 & 50.93 $\pm$ 3.70 & 51.85 $\pm$ 12.96 & 44.44 $\pm$ 12.96 & 0.00 $\pm$ 0.00 \\
Llama 2 & 70B & Y & 44.44 $\pm$ 12.96 & 46.30 $\pm$ 7.41 & 47.22 $\pm$ 4.17 & 44.44 $\pm$ 12.96 & 51.85 $\pm$ 12.96 & 29.63 $\pm$ 8.33 \\
Claude 2 & - & Y & - & - & - & - & - & 50.00 $\pm$ 7.41 \\
GPT 3.5 T & - & Y & - & 51.85 $\pm$ 5.09 & 46.30 $\pm$ 6.48 & 51.85 $\pm$ 12.96 & 55.56 $\pm$ 12.96 & 54.63 $\pm$ 9.26 \\
GPT 4 & - & Y & - & 56.48 $\pm$ 7.41 & 58.33 $\pm$ 9.26 & \textbf{64.81} $\pm$ \textbf{12.96} & 62.96 $\pm$ 12.96 & \textbf{71.30} $\pm$ \textbf{8.80} \\
\bottomrule
\end{tabular}

}
\caption{Results for the 55 example pairs with the Location of Element tag. Scores with 95\% confidence intervals above chance are shown in \textbf{bold}.}
\end{table*}
\begin{table*}
\centering
\resizebox{\textwidth}{!}{
\begin{tabular}{lrr|rrrrrr}
\toprule
Model & Params & Chat & Gen$^L$ & Val ZS$^L$ & Val FS$^L$ & Val ZS$^L$ (R) & Val FS$^L$ (R) & Val CoT$^T$ \\
\midrule
Llama 2 & 7B & N & \textbf{66.67} $\pm$ \textbf{14.29} & 50.00 $\pm$ 0.00 & 47.62 $\pm$ 4.76 & 52.38 $\pm$ 14.29 & 52.38 $\pm$ 14.29 & 7.14 $\pm$ 5.36 \\
Llama 2 & 7B & Y & 61.90 $\pm$ 14.29 & 54.76 $\pm$ 5.36 & 48.81 $\pm$ 1.79 & 54.76 $\pm$ 14.29 & 52.38 $\pm$ 14.29 & 25.00 $\pm$ 8.93 \\
Mistral 0.1 & 7B & N & \textbf{66.67} $\pm$ \textbf{14.29} & 50.00 $\pm$ 0.00 & 47.62 $\pm$ 2.98 & 52.38 $\pm$ 14.29 & 52.38 $\pm$ 14.29 & 0.00 $\pm$ 0.00 \\
Starling $\alpha$ & 7B & Y & \textbf{66.67} $\pm$ \textbf{14.29} & 53.57 $\pm$ 5.36 & 51.19 $\pm$ 3.57 & 59.52 $\pm$ 14.29 & 54.76 $\pm$ 14.29 & 32.14 $\pm$ 9.52 \\
Mistral 0.2 & 7B & Y & 57.14 $\pm$ 14.29 & 48.81 $\pm$ 8.33 & 54.76 $\pm$ 7.14 & 50.00 $\pm$ 14.29 & 50.00 $\pm$ 14.29 & 55.95 $\pm$ 8.33 \\
Llama 2 & 13B & N & \textbf{73.81} $\pm$ \textbf{13.10} & 45.24 $\pm$ 5.95 & 54.76 $\pm$ 5.36 & 52.38 $\pm$ 14.29 & 54.76 $\pm$ 14.29 & 3.57 $\pm$ 4.17 \\
Llama 2 & 13B & Y & \textbf{71.43} $\pm$ \textbf{13.10} & 55.95 $\pm$ 8.33 & 48.81 $\pm$ 1.79 & 59.52 $\pm$ 14.29 & 52.38 $\pm$ 14.29 & 13.10 $\pm$ 6.55 \\
Mixtral 0.1 & 8x7B & N & 57.14 $\pm$ 14.29 & 55.95 $\pm$ 7.14 & 46.43 $\pm$ 4.17 & 52.38 $\pm$ 14.29 & 54.76 $\pm$ 14.29 & 3.57 $\pm$ 4.17 \\
Mixtral 0.1 & 8x7B & Y & 57.14 $\pm$ 14.29 & 53.57 $\pm$ 7.14 & 52.38 $\pm$ 6.55 & 42.86 $\pm$ 14.29 & 54.76 $\pm$ 14.29 & 34.52 $\pm$ 8.33 \\
Llama 2 & 70B & N & \textbf{69.05} $\pm$ \textbf{14.29} & 52.38 $\pm$ 7.14 & \textbf{61.90} $\pm$ \textbf{6.55} & 61.90 $\pm$ 14.29 & 61.90 $\pm$ 14.29 & 2.38 $\pm$ 2.98 \\
Llama 2 & 70B & Y & \textbf{64.29} $\pm$ \textbf{14.29} & 55.95 $\pm$ 8.33 & 50.00 $\pm$ 3.57 & 59.52 $\pm$ 14.29 & 61.90 $\pm$ 14.29 & 13.10 $\pm$ 7.74 \\
Claude 2 & - & Y & - & - & - & - & - & \textbf{59.52} $\pm$ \textbf{8.33} \\
GPT 3.5 T & - & Y & - & 48.81 $\pm$ 4.76 & 53.57 $\pm$ 7.14 & \textbf{64.29} $\pm$ \textbf{14.29} & 61.90 $\pm$ 14.29 & 47.62 $\pm$ 9.52 \\
GPT 4 & - & Y & - & 50.00 $\pm$ 7.14 & \textbf{58.33} $\pm$ \textbf{7.14} & 59.52 $\pm$ 14.29 & 59.52 $\pm$ 14.29 & 55.95 $\pm$ 9.52 \\
\bottomrule
\end{tabular}

}
\caption{Results for the 42 example pairs with the Sensory tag. Scores with 95\% confidence intervals above chance are shown in \textbf{bold}.}
\end{table*}
\begin{table*}
\centering
\resizebox{\textwidth}{!}{
\begin{tabular}{lrr|rrrrrr}
\toprule
Model & Params & Chat & Gen$^L$ & Val ZS$^L$ & Val FS$^L$ & Val ZS$^L$ (R) & Val FS$^L$ (R) & Val CoT$^T$ \\
\midrule
Llama 2 & 7B & N & 62.50 $\pm$ 13.54 & 50.00 $\pm$ 0.00 & 48.96 $\pm$ 6.25 & 41.67 $\pm$ 14.58 & 41.67 $\pm$ 14.58 & 9.38 $\pm$ 5.21 \\
Llama 2 & 7B & Y & 58.33 $\pm$ 14.58 & 48.96 $\pm$ 6.25 & 50.00 $\pm$ 0.00 & 31.25 $\pm$ 12.50 & 62.50 $\pm$ 14.58 & 8.33 $\pm$ 5.21 \\
Mistral 0.1 & 7B & N & 62.50 $\pm$ 13.54 & 46.88 $\pm$ 4.17 & 48.96 $\pm$ 3.12 & 41.67 $\pm$ 13.54 & 43.75 $\pm$ 14.58 & 0.00 $\pm$ 0.00 \\
Starling $\alpha$ & 7B & Y & \textbf{64.58} $\pm$ \textbf{13.54} & 48.96 $\pm$ 3.12 & 48.96 $\pm$ 1.56 & 39.58 $\pm$ 13.54 & 52.08 $\pm$ 14.58 & 31.25 $\pm$ 8.85 \\
Mistral 0.2 & 7B & Y & \textbf{64.58} $\pm$ \textbf{13.54} & 45.83 $\pm$ 8.33 & 50.00 $\pm$ 6.25 & 47.92 $\pm$ 14.58 & 41.67 $\pm$ 14.58 & 46.88 $\pm$ 8.33 \\
Llama 2 & 13B & N & 60.42 $\pm$ 13.54 & 48.96 $\pm$ 6.25 & 47.92 $\pm$ 8.33 & 43.75 $\pm$ 14.58 & 47.92 $\pm$ 14.58 & 5.21 $\pm$ 4.17 \\
Llama 2 & 13B & Y & 56.25 $\pm$ 14.58 & 46.88 $\pm$ 7.29 & 48.96 $\pm$ 4.69 & 45.83 $\pm$ 14.58 & 39.58 $\pm$ 14.58 & 5.21 $\pm$ 4.17 \\
Mixtral 0.1 & 8x7B & N & 54.17 $\pm$ 14.58 & 52.08 $\pm$ 6.25 & 50.00 $\pm$ 0.00 & 33.33 $\pm$ 13.54 & 52.08 $\pm$ 14.58 & 2.08 $\pm$ 2.60 \\
Mixtral 0.1 & 8x7B & Y & 60.42 $\pm$ 14.58 & 52.08 $\pm$ 7.29 & 52.08 $\pm$ 6.25 & 50.00 $\pm$ 14.58 & 60.42 $\pm$ 14.58 & 46.88 $\pm$ 8.33 \\
Llama 2 & 70B & N & 60.42 $\pm$ 13.54 & 47.92 $\pm$ 6.25 & 53.12 $\pm$ 5.21 & 52.08 $\pm$ 14.58 & 47.92 $\pm$ 14.58 & 1.04 $\pm$ 1.56 \\
Llama 2 & 70B & Y & 54.17 $\pm$ 14.58 & 50.00 $\pm$ 7.29 & 47.92 $\pm$ 4.17 & 41.67 $\pm$ 14.58 & 56.25 $\pm$ 13.54 & 23.96 $\pm$ 8.33 \\
Claude 2 & - & Y & - & - & - & - & - & 57.29 $\pm$ 7.29 \\
GPT 3.5 T & - & Y & - & 52.08 $\pm$ 4.17 & 53.12 $\pm$ 6.25 & 54.17 $\pm$ 14.58 & 60.42 $\pm$ 13.54 & 45.83 $\pm$ 9.38 \\
GPT 4 & - & Y & - & \textbf{61.46} $\pm$ \textbf{6.77} & \textbf{62.50} $\pm$ \textbf{7.81} & \textbf{72.92} $\pm$ \textbf{12.50} & \textbf{70.83} $\pm$ \textbf{12.50} & \textbf{62.50} $\pm$ \textbf{9.38} \\
\bottomrule
\end{tabular}

}
\caption{Results for the 48 example pairs with the Sub-Word tag. Scores with 95\% confidence intervals above chance are shown in \textbf{bold}.}
\end{table*}
\begin{table*}
\centering
\resizebox{\textwidth}{!}{
\begin{tabular}{lrr|rrrrrr}
\toprule
Model & Params & Chat & Gen$^L$ & Val ZS$^L$ & Val FS$^L$ & Val ZS$^L$ (R) & Val FS$^L$ (R) & Val CoT$^T$ \\
\midrule
Llama 2 & 7B & N & 64.52 $\pm$ 16.13 & 51.61 $\pm$ 2.42 & 53.23 $\pm$ 6.45 & 54.84 $\pm$ 16.13 & \textbf{70.97} $\pm$ \textbf{16.13} & 4.84 $\pm$ 5.65 \\
Llama 2 & 7B & Y & 54.84 $\pm$ 16.13 & 56.45 $\pm$ 7.26 & 50.00 $\pm$ 0.00 & \textbf{67.74} $\pm$ \textbf{16.13} & 54.84 $\pm$ 16.13 & 11.29 $\pm$ 7.26 \\
Mistral 0.1 & 7B & N & 58.06 $\pm$ 16.13 & \textbf{58.06} $\pm$ \textbf{6.45} & 51.61 $\pm$ 2.42 & \textbf{67.74} $\pm$ \textbf{16.13} & \textbf{67.74} $\pm$ \textbf{16.13} & 0.00 $\pm$ 0.00 \\
Starling $\alpha$ & 7B & Y & 51.61 $\pm$ 16.13 & \textbf{62.90} $\pm$ \textbf{7.26} & 53.23 $\pm$ 4.03 & \textbf{67.74} $\pm$ \textbf{16.13} & 54.84 $\pm$ 17.74 & 46.77 $\pm$ 11.29 \\
Mistral 0.2 & 7B & Y & 61.29 $\pm$ 16.13 & 53.23 $\pm$ 12.90 & 54.84 $\pm$ 9.68 & 64.52 $\pm$ 16.13 & 61.29 $\pm$ 16.13 & 53.23 $\pm$ 11.29 \\
Llama 2 & 13B & N & \textbf{67.74} $\pm$ \textbf{16.13} & 59.68 $\pm$ 9.68 & 59.68 $\pm$ 10.48 & \textbf{67.74} $\pm$ \textbf{16.13} & \textbf{77.42} $\pm$ \textbf{14.52} & 8.06 $\pm$ 6.45 \\
Llama 2 & 13B & Y & 58.06 $\pm$ 16.13 & \textbf{62.90} $\pm$ \textbf{9.68} & \textbf{56.45} $\pm$ \textbf{5.65} & \textbf{67.74} $\pm$ \textbf{16.13} & 64.52 $\pm$ 16.13 & 8.06 $\pm$ 7.26 \\
Mixtral 0.1 & 8x7B & N & 58.06 $\pm$ 16.13 & \textbf{69.35} $\pm$ \textbf{9.68} & 54.84 $\pm$ 4.84 & \textbf{74.19} $\pm$ \textbf{16.13} & 61.29 $\pm$ 16.13 & 1.61 $\pm$ 2.42 \\
Mixtral 0.1 & 8x7B & Y & 54.84 $\pm$ 17.74 & 50.00 $\pm$ 9.68 & 54.84 $\pm$ 6.45 & 58.06 $\pm$ 16.13 & 38.71 $\pm$ 16.13 & 41.94 $\pm$ 9.68 \\
Llama 2 & 70B & N & 64.52 $\pm$ 16.13 & \textbf{66.13} $\pm$ \textbf{9.68} & \textbf{62.90} $\pm$ \textbf{7.26} & \textbf{67.74} $\pm$ \textbf{16.13} & \textbf{67.74} $\pm$ \textbf{16.13} & 1.61 $\pm$ 2.42 \\
Llama 2 & 70B & Y & 61.29 $\pm$ 17.74 & \textbf{67.74} $\pm$ \textbf{12.10} & 50.00 $\pm$ 4.84 & \textbf{70.97} $\pm$ \textbf{16.13} & \textbf{74.19} $\pm$ \textbf{14.52} & 25.81 $\pm$ 11.29 \\
Claude 2 & - & Y & - & - & - & - & - & 59.68 $\pm$ 9.68 \\
GPT 3.5 T & - & Y & - & 54.84 $\pm$ 5.65 & \textbf{64.52} $\pm$ \textbf{8.06} & 58.06 $\pm$ 16.13 & 64.52 $\pm$ 16.13 & 46.77 $\pm$ 12.10 \\
GPT 4 & - & Y & - & \textbf{64.52} $\pm$ \textbf{11.29} & 61.29 $\pm$ 11.29 & \textbf{67.74} $\pm$ \textbf{16.13} & \textbf{70.97} $\pm$ \textbf{16.13} & \textbf{75.81} $\pm$ \textbf{12.10} \\
\bottomrule
\end{tabular}

}
\caption{Results for the 31 example pairs with the Existence of Element tag. Scores with 95\% confidence intervals above chance are shown in \textbf{bold}.}
\end{table*}
\begin{table*}
\centering
\resizebox{\textwidth}{!}{
\begin{tabular}{lrr|rrrrrr}
\toprule
Model & Params & Chat & Gen$^L$ & Val ZS$^L$ & Val FS$^L$ & Val ZS$^L$ (R) & Val FS$^L$ (R) & Val CoT$^T$ \\
\midrule
Llama 2 & 7B & N & 58.33 $\pm$ 20.83 & 50.00 $\pm$ 0.00 & 50.00 $\pm$ 0.00 & 41.67 $\pm$ 20.83 & 41.67 $\pm$ 20.83 & 6.25 $\pm$ 6.25 \\
Llama 2 & 7B & Y & 58.33 $\pm$ 18.75 & 43.75 $\pm$ 6.25 & 50.00 $\pm$ 0.00 & 54.17 $\pm$ 20.83 & 58.33 $\pm$ 20.83 & 29.17 $\pm$ 10.42 \\
Mistral 0.1 & 7B & N & 41.67 $\pm$ 20.83 & 43.75 $\pm$ 7.29 & 52.08 $\pm$ 3.12 & 45.83 $\pm$ 20.83 & 54.17 $\pm$ 20.83 & 0.00 $\pm$ 0.00 \\
Starling $\alpha$ & 7B & Y & 37.50 $\pm$ 20.83 & 50.00 $\pm$ 6.25 & 47.92 $\pm$ 3.12 & 45.83 $\pm$ 20.83 & 58.33 $\pm$ 20.83 & 22.92 $\pm$ 12.50 \\
Mistral 0.2 & 7B & Y & 45.83 $\pm$ 20.83 & 50.00 $\pm$ 10.42 & 47.92 $\pm$ 8.33 & 41.67 $\pm$ 20.83 & 50.00 $\pm$ 20.83 & 41.67 $\pm$ 12.50 \\
Llama 2 & 13B & N & 58.33 $\pm$ 20.83 & 45.83 $\pm$ 5.21 & 54.17 $\pm$ 8.33 & 25.00 $\pm$ 16.67 & 50.00 $\pm$ 20.83 & 4.17 $\pm$ 5.21 \\
Llama 2 & 13B & Y & 54.17 $\pm$ 20.83 & 50.00 $\pm$ 8.33 & 47.92 $\pm$ 3.12 & 50.00 $\pm$ 20.83 & 41.67 $\pm$ 20.83 & 16.67 $\pm$ 9.38 \\
Mixtral 0.1 & 8x7B & N & 50.00 $\pm$ 20.83 & 50.00 $\pm$ 8.33 & 47.92 $\pm$ 3.12 & 45.83 $\pm$ 20.83 & 50.00 $\pm$ 20.83 & 2.08 $\pm$ 3.12 \\
Mixtral 0.1 & 8x7B & Y & 41.67 $\pm$ 20.83 & 43.75 $\pm$ 8.33 & 47.92 $\pm$ 10.42 & 37.50 $\pm$ 18.75 & 37.50 $\pm$ 18.85 & 29.17 $\pm$ 12.50 \\
Llama 2 & 70B & N & 50.00 $\pm$ 20.83 & 45.83 $\pm$ 8.33 & 54.17 $\pm$ 10.42 & 41.67 $\pm$ 20.83 & 45.83 $\pm$ 20.83 & 4.17 $\pm$ 5.21 \\
Llama 2 & 70B & Y & 54.17 $\pm$ 20.83 & 52.08 $\pm$ 7.29 & 45.83 $\pm$ 5.21 & 41.67 $\pm$ 20.83 & 54.17 $\pm$ 20.83 & 22.92 $\pm$ 10.42 \\
Claude 2 & - & Y & - & - & - & - & - & 41.67 $\pm$ 10.42 \\
GPT 3.5 T & - & Y & - & 54.17 $\pm$ 5.21 & 45.83 $\pm$ 8.33 & \textbf{70.83} $\pm$ \textbf{16.67} & 54.17 $\pm$ 20.83 & 50.00 $\pm$ 10.42 \\
GPT 4 & - & Y & - & 47.92 $\pm$ 9.38 & 52.08 $\pm$ 14.58 & 41.67 $\pm$ 20.83 & 62.50 $\pm$ 18.75 & 45.83 $\pm$ 12.50 \\
\bottomrule
\end{tabular}

}
\caption{Results for the 24 example pairs with the Hypothetical tag. Scores with 95\% confidence intervals above chance are shown in \textbf{bold}.}
\end{table*}
\begin{table*}
\centering
\resizebox{\textwidth}{!}{
\begin{tabular}{lrr|rrrrrr}
\toprule
Model & Params & Chat & Gen$^L$ & Val ZS$^L$ & Val FS$^L$ & Val ZS$^L$ (R) & Val FS$^L$ (R) & Val CoT$^T$ \\
\midrule
Llama 2 & 7B & N & 44.00 $\pm$ 20.00 & 50.00 $\pm$ 6.00 & 50.00 $\pm$ 6.00 & 40.00 $\pm$ 20.00 & 56.00 $\pm$ 20.00 & 2.00 $\pm$ 3.00 \\
Llama 2 & 7B & Y & 60.00 $\pm$ 20.00 & 50.00 $\pm$ 10.00 & 50.00 $\pm$ 0.00 & 52.00 $\pm$ 20.00 & 56.00 $\pm$ 20.00 & 6.00 $\pm$ 6.00 \\
Mistral 0.1 & 7B & N & 64.00 $\pm$ 18.00 & 50.00 $\pm$ 8.00 & 46.00 $\pm$ 5.00 & 52.00 $\pm$ 20.00 & 60.00 $\pm$ 20.00 & 0.00 $\pm$ 0.00 \\
Starling $\alpha$ & 7B & Y & 64.00 $\pm$ 20.00 & 54.00 $\pm$ 10.00 & 50.00 $\pm$ 0.00 & 56.00 $\pm$ 20.00 & 60.00 $\pm$ 20.00 & 44.00 $\pm$ 15.00 \\
Mistral 0.2 & 7B & Y & 48.00 $\pm$ 20.00 & 56.00 $\pm$ 12.00 & 58.00 $\pm$ 10.00 & 60.00 $\pm$ 20.00 & 60.00 $\pm$ 20.00 & 58.00 $\pm$ 12.00 \\
Llama 2 & 13B & N & 60.00 $\pm$ 20.00 & 56.00 $\pm$ 8.00 & 50.00 $\pm$ 12.00 & 52.00 $\pm$ 20.00 & \textbf{68.00} $\pm$ \textbf{18.00} & 6.00 $\pm$ 7.00 \\
Llama 2 & 13B & Y & 56.00 $\pm$ 20.00 & 54.00 $\pm$ 12.00 & 54.00 $\pm$ 8.00 & 52.00 $\pm$ 20.00 & 52.00 $\pm$ 20.00 & 8.00 $\pm$ 9.00 \\
Mixtral 0.1 & 8x7B & N & 60.00 $\pm$ 20.00 & 58.00 $\pm$ 10.00 & 54.00 $\pm$ 8.00 & 64.00 $\pm$ 20.00 & 52.00 $\pm$ 20.00 & 2.00 $\pm$ 3.00 \\
Mixtral 0.1 & 8x7B & Y & 60.00 $\pm$ 20.00 & 44.00 $\pm$ 12.00 & 54.00 $\pm$ 5.00 & 60.00 $\pm$ 20.00 & 56.00 $\pm$ 20.00 & 52.00 $\pm$ 16.00 \\
Llama 2 & 70B & N & 56.00 $\pm$ 20.00 & \textbf{60.00} $\pm$ \textbf{8.00} & 56.00 $\pm$ 6.00 & 56.00 $\pm$ 20.00 & 56.00 $\pm$ 20.00 & 0.00 $\pm$ 0.00 \\
Llama 2 & 70B & Y & 56.00 $\pm$ 20.00 & 52.00 $\pm$ 10.00 & 48.00 $\pm$ 3.00 & 56.00 $\pm$ 20.00 & 56.00 $\pm$ 20.00 & 30.00 $\pm$ 10.00 \\
Claude 2 & - & Y & - & - & - & - & - & \textbf{68.00} $\pm$ \textbf{10.00} \\
GPT 3.5 T & - & Y & - & 56.00 $\pm$ 8.00 & 54.00 $\pm$ 8.00 & 56.00 $\pm$ 20.00 & \textbf{72.00} $\pm$ \textbf{18.00} & 42.00 $\pm$ 13.05 \\
GPT 4 & - & Y & - & \textbf{68.00} $\pm$ \textbf{13.00} & \textbf{64.00} $\pm$ \textbf{13.00} & \textbf{84.00} $\pm$ \textbf{14.00} & \textbf{80.00} $\pm$ \textbf{16.00} & \textbf{80.00} $\pm$ \textbf{11.00} \\
\bottomrule
\end{tabular}

}
\caption{Results for the 25 example pairs with the Grammaticality tag. Scores with 95\% confidence intervals above chance are shown in \textbf{bold}.}
\end{table*}
\begin{table*}
\centering
\resizebox{\textwidth}{!}{
\begin{tabular}{lrr|rrrrrr}
\toprule
Model & Params & Chat & Gen$^L$ & Val ZS$^L$ & Val FS$^L$ & Val ZS$^L$ (R) & Val FS$^L$ (R) & Val CoT$^T$ \\
\midrule
Llama 2 & 7B & N & 59.09 $\pm$ 20.45 & 52.27 $\pm$ 6.82 & 59.09 $\pm$ 10.23 & 59.09 $\pm$ 18.18 & 54.55 $\pm$ 20.45 & 2.27 $\pm$ 3.41 \\
Llama 2 & 7B & Y & 59.09 $\pm$ 20.45 & 63.64 $\pm$ 13.64 & 52.27 $\pm$ 3.41 & 59.09 $\pm$ 18.18 & 54.55 $\pm$ 22.73 & 20.45 $\pm$ 10.23 \\
Mistral 0.1 & 7B & N & 36.36 $\pm$ 20.45 & \textbf{63.64} $\pm$ \textbf{11.36} & 50.00 $\pm$ 0.00 & \textbf{86.36} $\pm$ \textbf{13.64} & 59.09 $\pm$ 20.45 & 0.00 $\pm$ 0.00 \\
Starling $\alpha$ & 7B & Y & 50.00 $\pm$ 20.45 & \textbf{63.64} $\pm$ \textbf{11.36} & 52.27 $\pm$ 6.82 & 63.64 $\pm$ 20.45 & 54.55 $\pm$ 22.73 & 38.64 $\pm$ 13.64 \\
Mistral 0.2 & 7B & Y & 54.55 $\pm$ 22.73 & 59.09 $\pm$ 14.77 & 50.00 $\pm$ 11.36 & 63.64 $\pm$ 20.45 & 68.18 $\pm$ 18.18 & 43.18 $\pm$ 12.50 \\
Llama 2 & 13B & N & 59.09 $\pm$ 20.45 & 61.36 $\pm$ 13.64 & \textbf{70.45} $\pm$ \textbf{13.64} & 68.18 $\pm$ 18.18 & \textbf{77.27} $\pm$ \textbf{18.18} & 2.27 $\pm$ 3.41 \\
Llama 2 & 13B & Y & 63.64 $\pm$ 20.45 & 56.82 $\pm$ 13.64 & 59.09 $\pm$ 11.36 & 59.09 $\pm$ 20.45 & 63.64 $\pm$ 18.18 & 13.64 $\pm$ 12.50 \\
Mixtral 0.1 & 8x7B & N & 59.09 $\pm$ 20.45 & \textbf{72.73} $\pm$ \textbf{14.77} & \textbf{65.91} $\pm$ \textbf{9.09} & \textbf{72.73} $\pm$ \textbf{18.18} & 68.18 $\pm$ 18.18 & 0.00 $\pm$ 0.00 \\
Mixtral 0.1 & 8x7B & Y & 45.45 $\pm$ 20.45 & 61.36 $\pm$ 15.91 & \textbf{65.91} $\pm$ \textbf{9.09} & \textbf{72.73} $\pm$ \textbf{18.18} & 68.18 $\pm$ 18.18 & 52.27 $\pm$ 13.64 \\
Llama 2 & 70B & N & 63.64 $\pm$ 20.45 & 59.09 $\pm$ 13.64 & 54.55 $\pm$ 11.36 & \textbf{72.73} $\pm$ \textbf{18.18} & \textbf{72.73} $\pm$ \textbf{18.18} & 0.00 $\pm$ 0.00 \\
Llama 2 & 70B & Y & 63.64 $\pm$ 20.45 & \textbf{68.18} $\pm$ \textbf{14.77} & 56.82 $\pm$ 9.09 & \textbf{72.73} $\pm$ \textbf{18.18} & 68.18 $\pm$ 18.18 & 9.09 $\pm$ 10.23 \\
Claude 2 & - & Y & - & - & - & - & - & 47.73 $\pm$ 13.64 \\
GPT 3.5 T & - & Y & - & \textbf{65.91} $\pm$ \textbf{13.64} & 63.64 $\pm$ 13.64 & 59.09 $\pm$ 20.45 & \textbf{72.73} $\pm$ \textbf{18.18} & \textbf{63.64} $\pm$ \textbf{11.36} \\
GPT 4 & - & Y & - & 61.36 $\pm$ 11.36 & 59.09 $\pm$ 13.64 & \textbf{86.36} $\pm$ \textbf{13.64} & \textbf{81.82} $\pm$ \textbf{15.91} & \textbf{72.73} $\pm$ \textbf{10.23} \\
\bottomrule
\end{tabular}

}
\caption{Results for the 22 example pairs with the Question tag. Scores with 95\% confidence intervals above chance are shown in \textbf{bold}.}
\end{table*}

\end{document}